\documentclass[10pt,twocolumn,letterpaper]{article}

\usepackage{cvpr}              
\usepackage{graphicx}
\usepackage{amsmath}
\usepackage{amssymb}
\usepackage{booktabs}
\usepackage{multirow}
\usepackage{makecell}
\usepackage{amsthm,amsmath,amssymb}
\usepackage{mathrsfs}
\usepackage[pagebackref,breaklinks,colorlinks]{hyperref}
\usepackage{float}
\usepackage{pifont}
\usepackage{graphicx}
\usepackage{epstopdf}
\usepackage{caption}
\usepackage{tabularx}
\usepackage{subcaption, enumitem}
\newtheorem{problem}{Problem}
\newtheorem{theorem}{Theorem}[]
\newtheorem{proposition}{Remark}[]

\newtheorem{lemma}{Lemma}[]
\theoremstyle{definition}
\newtheorem{definition}{Definition}[]

\usepackage[capitalize]{cleveref}
\usepackage[dvipsnames]{xcolor}

\definecolor{cvprblue}{rgb}{0.21,0.49,0.74}

\def\paperID{5432} 
\def\confName{CVPR}
\def\confYear{2024}

\title{PNeRV: Enhancing Spatial Consistency via\\ Pyramidal Neural Representation for Videos}

\author{Qi Zhao\\
Nanjing University\\
{\tt\small qizhao@smail.nju.edu.cn}
\and
M. Salman Asif\\
University of California Riverside\\
{\tt\small sasif@ucr.edu}
\and
Zhan Ma\footnote[1]\\ \\
Nanjing University\\
{\tt\small mazhan@nju.edu.cn}
}
\vspace{-0.05in}

\begin{document}
\twocolumn[{%
\renewcommand\twocolumn[1][]{#1}%
\maketitle
\begin{center}
  \centering
   \setlength\belowdisplayskip{-0.1in}
   \includegraphics[width=0.85\linewidth]{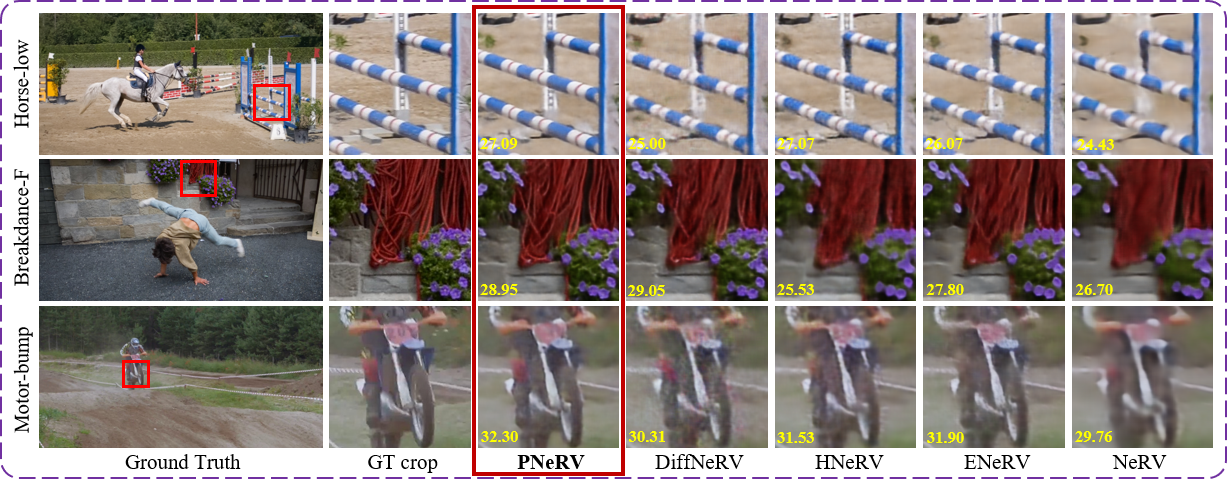}
    \setlength{\belowcaptionskip}{0.05in}
    \setlength{\abovecaptionskip}{0.05in}
   \captionof{figure}{High-quality video ($1920 \times 960$) reconstruction comparisons between the proposed \textbf{P}yramidal \textbf{NeRV} and other models, PSNR in \textcolor{yellow}{yellow}. PNeRV outperforms other models on perceptual quality with less noise and artifacts, maintaining spatial consistency.}
   \label{fig1}
\end{center}%
}]
\footnotetext[1]{Corresponding author: Zhan Ma (\tt\small mazhan@nju.edu.cn)}

\begin{abstract}
The primary focus of Neural Representation for Videos (NeRV) is to effectively model its spatiotemporal consistency. However, current NeRV systems often face a significant issue of spatial inconsistency, leading to decreased perceptual quality. To address this issue, we introduce the Pyramidal Neural Representation for Videos (PNeRV), which is built on a multi-scale information connection and comprises a lightweight rescaling operator, Kronecker Fully-connected layer (KFc), and a Benign Selective Memory (BSM) mechanism.
The KFc, inspired by the tensor decomposition of the vanilla Fully-connected layer, facilitates low-cost rescaling and global correlation modeling. BSM merges high-level features with granular ones adaptively. Furthermore, we provide an analysis based on the Universal Approximation Theory of the NeRV system and validate the effectiveness of the proposed PNeRV.
We conducted comprehensive experiments to demonstrate that PNeRV surpasses the performance of contemporary NeRV models, achieving the best results in video regression on UVG and DAVIS under various metrics (PSNR, SSIM, LPIPS, and FVD). Compared to vanilla NeRV, PNeRV achieves a +4.49 dB gain in PSNR and a 231\% increase in FVD on UVG, along with a +3.28 dB PSNR and 634\% FVD increase on DAVIS.
\end{abstract}  \vspace{-0.1in}  
\section{Introduction}
\label{sec:intro}

In recent years, Implicit Neural Representation (INR) has emerged as a pivotal area of research across various vision tasks, including neural radiance fields modeling~\cite{nerf, pixelNeRFNR}, 3D vision~\cite{GRAFGR, EfficientG3, NeuralGL} and multimedia neural coding~\cite{siren, nerv}. INR operates on the philosophy that target implicit mapping will be encoded into a learnable neural network through end-to-end training. By leveraging the modeling capabilities of neural nets, INR can approximate a wide range of complex nonlinear or high-dimensional mappings.

However, when considering the video coding task, extant NeRV systems exhibit a notable deficiency in perceptual quality. The reconstructions of foreground subjects, which are obscured by high-frequency irrelevant details or blurring, prove challenging for current NeRV models. This issue of spatial inconsistency is primarily attributed to semantic uncertainty, causing the model to struggle with discerning whether two long-range pixels pertain to the same objects or constitute part of a noisy background. We postulate that this predicament stems from the absence of \textbf{global receptive field} and \textbf{multi-scale information communication}. Inspired by existing empirical evidence from other vision research, we speculate that if the dense prediction could leverage the high-level information learned from raw input, it would substantially alleviate both the semantic uncertainty and spatial inconsistency (as illustrated in Fig.~\ref{fig1}).

In practice, introducing multi-scale structures into NeRV poses a significant and non-trivial challenge. Existing NeRV models typically resort to cascaded upsampling layers (the so-called “mainstream”) for decoding fine video, striking a compromise between performance and efficiency. However, layers that use subpixel-based operators~\cite{ps, deconv} can hardly maintain a balance between the increasing receptive field, parameter demand, and performance (more discussions in Sec.~\ref{supp_rw} and visualization in Fig.~\ref{ps_kfc}). Additionally, these decoding layers are solely receptive to features from the previous layer, ignoring information from other preceding layers. Moreover, the design of multi-scale structures in NeRV remains unguided by either practical or theoretical principles due to constraints on parameter quantities compared with methods for other vision tasks.

To address this issue, we propose the \textbf{P}yramidal \textbf{Ne}ural \textbf{R}epresentation for \textbf{V}ideos (\textbf{PNeRV}) based on hierarchical information interaction via a low-cost upscaling operator, \textit{Kronecker Fully-connected} (\textsc{KFc}) layer, and a gated mechanism, \textit{Benign Selective Memory} (BSM), which aims at adaptive feature merging. Utilizing these modules, PNeRV can fuse the high-level features directly into each underlying fine-grained layer via shortcuts, thereby creating a pyramidal structure. Further, we introduce \textit{Universal Approximation Theory} (UAT) into the NeRV system for the first time and provide an analysis of existing NeRV models, revealing the superiority of our proposed pyramid structure. Our main contributions are summarized as follows.

\begin{itemize}
\item Towards the poor perceptual quality of NeRV systems, we propose PNeRV to enhance spatial consistency via multi-scale feature learning.

\item In pursuit of model efficiency pursuit, we propose the \textsc{KFc}, which realizes low-cost upsampling with a global receptive field and BSM for adaptive feature fusion, thus forming an efficient multi-level pyramidal structure.

\item We introduce the first UAT analysis in NeRV research. Using UAT, we describe NeRV-based video neural coding as the Implicit Video Neural Coding problem, clarifying and defining some fundamental concepts within this framework.

\item We confirm the superiority of PNeRV against other models on two datasets (UVG and DAVIS) using four video quality metrics (PSNR, SSIM, LPIPS, and FVD).
\end{itemize}
\vspace{-0.02in}
\section{Related Work}
\vspace{-0.02in}
\label{sec:formatting}

\noindent\textbf{Implicit Neural Representation for Videos}. In recent years, INR has gained increasing attention in various vision areas, such as neural radiance fields modeling~\cite{nerf, instantNGP, mobilenerf, Plenoxels}, novel view synthesis~\cite{LLFF, MINE}, and multimedia neural coding~\cite{COIN, nerv, HNeRV, enerv, DNeRV}. 
For INR-based neural video coding, NeRV~\cite{nerv} first uses index embeddings as input and then decodes back to high-resolution videos via cascaded PixelShuffle~\cite{ps} blocks. ENeRV~\cite{enerv} aims to reallocate the parameter quantity between different modules for better performance. Unlike the above index-based methods, HNeRV~\cite{HNeRV} employs ConvNeXT~\cite{convnext} blocks as an encoder and provides \textit{content-aware embeddings}, improving the performance. Furthermore, apart from content embedding, DiffNeRV~\cite{DNeRV} inputs the difference between adjacent frames as \textit{temporal embeddings}, enhancing temporal consistency. The major distinction between PNeRV and DiffNeRV is that the latter does not refer to multi-scale spatial information, resulting in spatial discontinuity.

\noindent\textbf{Multi-scale Hierarchy Structure for Dense Prediction}. In previous CV research, there have emerged numerous studies on multi-scale vision~\cite{TheLP, UNet, SpatialPP, FPN, UNet++, yolov3, PANet, PVT, SwinTH}. UNet~\cite{UNet} aimed to improve accuracy by combining contextual information from features at different resolutions. FPN~\cite{FPN} developed a top-down architecture with high-level semantic feature maps at all scales, showing significant improvements in dense prediction tasks. PANet~\cite{PANet} followed the idea of multi-level information fusion and proposed adaptive feature pooling to leverage useful information from each level. PVT~\cite{PVT} introduced the pyramidal architecture into vision transformers. The success of pyramidal structure lies in multi-level feature fusion, and detailed predictions should be guided by high-level context features. 

\noindent\textbf{Video Coding Pipelines and Theories}. Video coding has been studied for several decades based on handcrafted design and domain transformation~\cite{mpeg, h264, h265, dct}. Furthermore, \textit{neural video coding}~\cite{dvc, dcvc, FVC, MLVC} aims to replace some components in the traditional pipeline, but they suffer from high computational complexity and slow decoding speeds. Beyond Rate-Distortion Optimization (RDO)~\cite{rdo}, \cite{Blau2017ThePT} reveals the importance of perceptual quality and proposes the Perception-Distortion Optimization (PDO). \cite{Blau2019RethinkingLC} defines the Rate-Distortion-Perception Optimization (RDPO). 
Different from those pipelines, we reinterpret the INR-based video coding~\cite{nerv, HNeRV, DNeRV} with UAT framework, and more details are in Sec.~\ref{invc_sec} and Sec.~{\color{red}A.1}.

\noindent\textbf{Universal Approximation Theory (UAT)}. One of the pursuits of UAT analysis on the deep neural net is to estimate the minimal width of a model to approximate continuous functions under certain errors and fixed lengths. \cite{ApproximatingCF} provides the estimation of minimal width $w^*$ of a \textsc{ReLU} net as $d_{in}+1 \leq w^* \leq d_{in}+d_{out}$ in Theorem~1. \cite{MinimumWF} provides the first definitive result for deep ReLU nets, and the minimum width required for the universal approximation of the $L^p$ functions is exactly $\mathop{\max}\{d_{in}+1,d_{out}\}$. \cite{ResNetWO} demonstrates that a deep ReLU ResNet with one neuron per hidden layer can uniformly approximate any Lebesgue integrable function. More discussions are given in Sec.~\ref{uat_sec} and Sec.~{\color{red}A.1}.

\vspace{-0.02in}
\section{Pyramidal Neural Representation for Videos}
\vspace{-0.02in}

\begin{figure}[!t]
  \centering
  \includegraphics[width=0.9\linewidth]{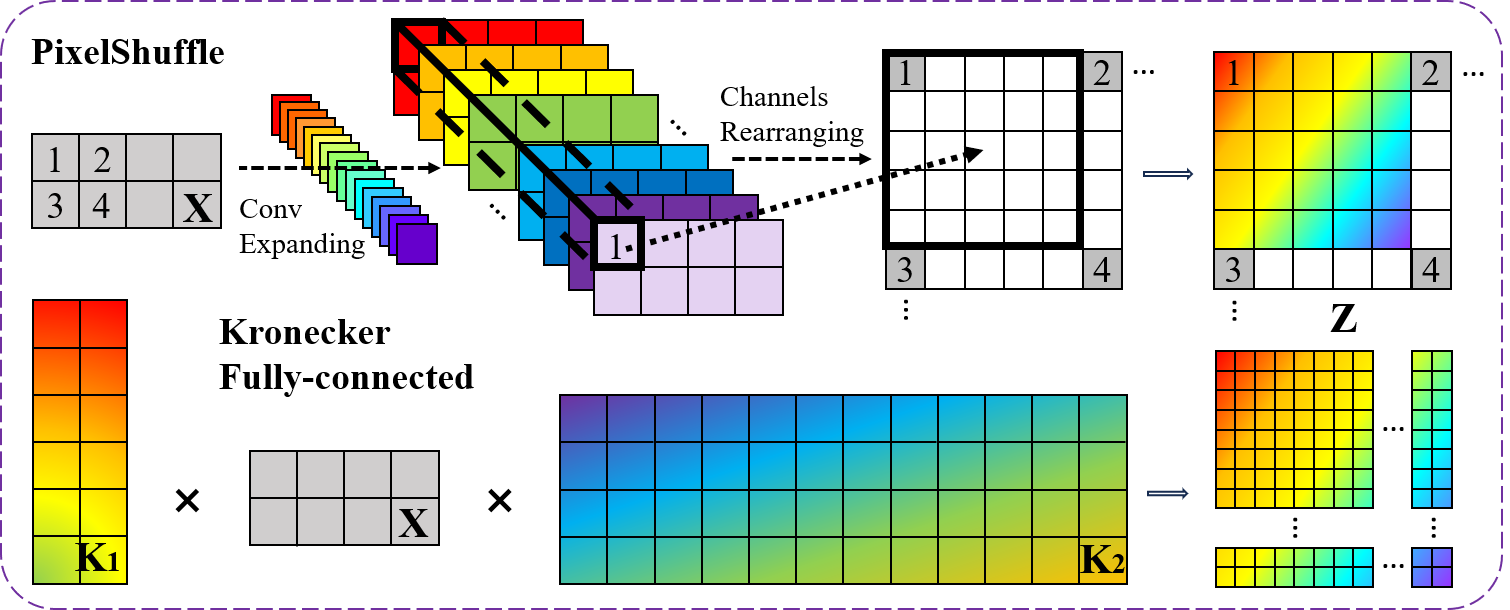}\\
  \vspace{-0.05in}\caption{Visualized comparison between PixelShuffle and KFc, where $\times$ denotes matrix multiplication and black box is the subpixel area. PixelShuffle fills the subpixels using a local receptive field, lacking long-range relationship modeling ability, while \textsc{KFc} calculates the correlation between every position.}
  \label{ps_kfc}\vspace{-0.2in}
\end{figure}

As analyzed before, pursuing spatial consistency leads to the communication of multi-scale information via a global receptive field. Fine-grained reconstruction requires high-level information as guidance and a low-cost upsampling operator is crucial for creating multi-level shortcuts.

Therefore, we propose \textbf{P}yramidal \textbf{NeRV} (\textbf{PNeRV}) consisting of a learnable encoder and a novel pyramidal decoder. The main innovation in the decoder is a low-cost global-wise upscaling operator, Kronecker Fully-connected (\textsc{KFc}) layer, and a gated memory unit, Benign Selective Memory (BSM) for disentangled feature fusion. The overall structure of PNeRV is shown in Fig.~\ref{pnerv}.

\vspace{-0.02in}
\subsection{Kronecker Fully-connected Layer}
\vspace{-0.02in}

NeRV aims to decode high-resolution videos from tiny embeddings. Therefore, Conv-based upsampling operators~\cite{deconv, ps} are not efficient enough due to the huge upscaling ratio, which differs from previous visual tasks. The parameter quantity will grow sharply due to increased channels or kernel size. 
However, NeRV aims to encode videos with as few parameters as possible, namely \textit{model efficiency pursuit}.

In contrast to this goal, subpixel-based upscaling operators fail to form shortcuts and a pyramidal structure. Once upscaling from given embeddings $F_0$ ($16\times2\times4$) to fine-grained features $F_n$ ($16\times320\times640$), there is an intolerable increase in parameters (25600$\times$) to fill in the target subpixels. Even when the kernel size is only $1\times1$, a single PixelShuffle~\cite{ps} layer requires 6.96M parameters from $F_0$ to $F_n$, regardless of the size of videos or model structure.

Towards this dilemma, we propose the Kronecker Fully-connected layer~(\textsc{KFc}), given as
\begin{equation}\label{kfc}\small
\mathbf{Z} = \mathop{\textsc{concat}}_i \left( \mathbf{K}_1^{(i)} \mathbf{X}^{(i)} \mathbf{K}_2^{(i)} \right) + \mathbf{b}_c \otimes \mathbf{b}_h \otimes \mathbf{b}_w,
\end{equation}
where $\mathbf{X}^{(i)} \in \mathbb{R}^{ H_{in} \times W_{in}}$ are input features, $\mathbf{Z}^{(i)} \in \mathbb{R}^{H_{out} \times W_{out}}$ are output features, $\mathbf{K}_{1,2}$ are two kernels which $\mathbf{K}_{1}^{(i)} \in \mathbb{R}^{H_{out} \times H_{in}}$ and $\mathbf{K}_{2}^{(i)} \in \mathbb{R}^{ W_{in} \times W_{out}}$ in channel $i$. Each feature map is calculated channel-wise and will be concatenated in the channel. $\mathbf{b}_{c,h,w}$ are three vectors and they output the \textsc{bias} via kronecker product $\otimes$ where $\mathbf{b}_{c} \in \mathbb{R}^{C \times 1}$, $\mathbf{b}_{h} \in \mathbb{R}^{H_{out}  \times 1}$ and $\mathbf{b}_{w} \in \mathbb{R}^{W_{out}  \times 1}$.

\noindent\textbf{Motivation}. \textsc{KFc} is motivated by the fact that, \textit{the subpixels of one position are related to every other position in current feature maps}. 
The dilemma between local and global feature learning is an enduring issue in deep learning~\cite{NonlocalNN, MultiScaleCA, AttentionIA, SwinTH}. 
Unlike the local prior in the \textsc{Conv} layer, \textsc{Fc} is more effective, especially for the top embeddings containing semantic features with little local spatial structure. The calculation between $\mathbf{K}_1$, $\mathbf{X}$ and $\mathbf{K}_2$ is actually the product between vectorized input features $\text{vec}(\mathbf{X}) \in \mathbb{R}^{ H_{in} W_{in}  \times 1}$ and hybrid weight matrix $\mathbf{K}_{\otimes} \in \mathbb{R}^{H_{out}W_{out} \times H_{in}W_{in}}$, where $\mathbf{K}_{\otimes} = \mathbf{K}_1 \otimes \mathbf{K}^{\top}_2$. Compared with the vanilla \textsc{Fc} layer, two low-rank matrices $\mathbf{K}_1$ and $\mathbf{K}_2$ come from the Kronecker decomposition, while the bias term $\mathbf{b}_{c,h,w}$ is the CP decomposition of the original ones. 

\begin{figure*}[!h]
  \centering
  \setlength\belowdisplayskip{-0.1in}
  \setlength{\abovecaptionskip}{0.05in}
  \includegraphics[width=0.85\linewidth]{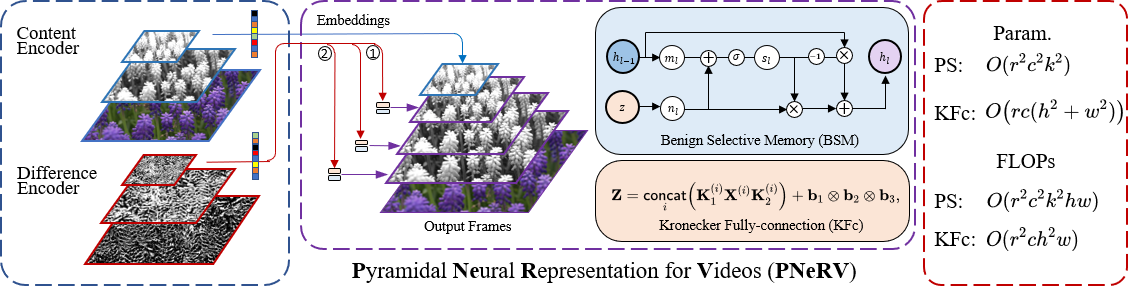}\\
  \vspace{-0.05in}\caption{The overall architecture of PNeRV, consists of \textsc{KFc} and BSM. The right part shows the comparison of parameters and FLOPs between PixelShuffle (PS) and \textsc{KFc}, where input feature maps are in $c \times h \times w$, the upscaling rate is $r$ and kernel size in PS is $k \times k$. 
  }
  \label{pnerv}\vspace{-0.1in}
\end{figure*}

Besides, \textsc{KFc} is also inspired by LoRA~\cite{lora}, which uses adaptive weights in low ``intrinsic dimension''~\cite{intD} for PEFT. Visualization is shown in Fig.~\ref{ps_kfc}. For the same $F_0$ and $F_n$ mentioned above, parameters needed by \textsc{KFc} is 0.05M, only $\textbf{0.7\%}$ of that required by PixelShuffle. Detailed comparisons of parameters and FLOPs are given in Fig.~\ref{pnerv}.

\vspace{-0.02in}
\subsection{Benign Selective Memory}
\vspace{-0.02in}
Using \textsc{KFc} as the basic operator for shortcuts, PNeRV realizes efficient multi-scale feature learning. Also, adaptive feature fusion between different levels is quite important. 

Therefore, we propose the Benign Selective Memory (BSM). BSM is inspired by the gated mechanism in RNN research~\cite{gru, LSTM}, treating features in different streams as input and cell states. We follow the convention in RNN, where lowercase represents hidden states. For the high-level feature $z$ on the top and the fine-grained feature $h_{l-1}$ in the $l$-th layer, BSM is given as follows:
\begin{small}
\begin{align*}
n_l =&~ W_n \ast z  ,       & \textsc{Kownledge} \\
m_l =&~ W_m \ast h_{l-1}        & \textsc{Memory}\\
s_l =&~ \mathop{ \sigma} (W_s \ast \textsc{ReLU} (n_l + m_l)  )  ,      &  \textsc{Decision}\\
h_{l} =&~ h_{l-1} \odot (1-s_l) +  n_l \odot s_l, &  \textsc{Behaviour}  \\[-0.2in] 
\end{align*}
\end{small}where $\ast$ is convolution with weights $W_{n,m,s}$ ,$\odot$ is hadamard product and $\sigma$ is the sigmoid activation.

BSM is an imitation of the human learning and decision-making process. The high-level $z$ is regarded as external Knowledge, while $h_{l-1}$ from the previous block in the mainstream is the inheriting Memory. The model should learn from Knowledge and integrate it with Memory to guide the Behaviors (reconstruction). That is the so-called Benign Selective Memory.

\noindent\textbf{Motivation}. The primary distinction between previous gated mechanisms and BSM is that BSM learns features (referred to as ``Knowledge” and ``Memory") separately before merging them. This disentangled fashion aids PNeRV in adaptively merging features from different levels. The ablation studies in Tab.~\ref{abla_da} show the superiority of BSM.

\vspace{-0.02in}
\subsection{Overall Structure}
\vspace{-0.02in}
Therefore, the proposed PNeRV consists of three parts, as follows (where $X$ is the input embedding, $\hat{H_l}$ are featured in the mainstream $l^{th}$ layer, $Z_l$ are features upsampled by shortcuts, and $H_l$ are the features after fusion):
\begin{enumerate}
\setlength\abovedisplayskip{-0.02in}
\setlength\belowdisplayskip{-0.02in}
\item A mainstream comprises cascaded upsampling layers (containing \textsc{Conv}, PixelShuffle, and \textsc{GELU}) to provide high-resolution reconstruction, $\hat{H}_l = Block(H_{l-1}), 1\leq l \leq L, L=6, H_0=X$.
\item Various shortcuts upsample the high-level embeddings $X$ into $Z_l$ before merging into the mainstream, forming a multi-level hierarchical architecture, $Z_l = Shortcut(X), 2\leq l \leq L_0, L_0=5$.
\item A feature fusion mechanism is employed to merge $Z_l$ with $\hat{H}_l$ adaptively for the final output, $H_l = Fusion(Z_l, \hat{H}_l)$.
\end{enumerate}

In implementation, we conducted two versions, namely PNeRV-M and PNeRV-L. PNeRV-M has only a single stream which takes content embeddings~\cite{HNeRV} $\mathbf{X}^C$ in $16 \times 2 \times 4$ as input.
For PNeRV-L, temporal embeddings~\cite{DNeRV} $\mathbf{X}^T$ in ${2 \times 40 \times 80}$ are involved. 
$\mathbf{X}^C$ is delivered to the mainstream and $\mathbf{X}^T$ is upscaled in shortcuts via \textsc{KFc} and merged into each mainstream layers through BSM. We choose PNeRV-L as the final version. All kernels are $3 \times 3$ except for the first and final output layer. For the input video $\mathbf{V}$ and reconstructions $\tilde{\mathbf{V}}$, the key equations of the entire model in the $l$-th layer ($1< l\leq L$) are presented as follows:
\vspace{-0.05in}
\begin{equation*}
\small
\begin{split}
\setlength\abovedisplayskip{-0.02in}
\setlength\belowdisplayskip{-0.05in}
 Encoder : ~ \mathbf{X}^C, \mathbf{X}^T =&~ \mathcal{E} ( \mathbf{V} ),   \\
 Decoder : ~~~~~~~~~~~\hat{\mathbf{H}}_{l} =&~ \textsc{Block}_l(\mathbf{H}_{l-1})\\
  =&~ \textsc{Block}_l \circ \cdots \circ \textsc{Block}_1 (\mathbf{X}^C ),  \\
  \mathbf{Z}_l =&~ \textsc{Shortcut}_l (\mathbf{X}^T ),        \\
\mathbf{H}_{l}  =&~  \textsc{BSM}_l( \hat{\mathbf{H}}_{l}, \mathbf{Z}_l ),     \\
\end{split}
\end{equation*}
where $\mathbf{H}_0 = \mathbf{X}^C$. The final output will be passed through an output layer, $\tilde{\mathbf{V}}=\textsc{Conv}_{1\times1}(\mathbf{H}_{l=L})$.

\section{Universal Approximation Theory on NeRV}\label{uat_sec}
First, we will clarify some concepts in NeRV within the UAT framework. A NeRV-based neural video coding pipeline is defined in Sec.~\ref{invc_sec}. We describe the limitations of existing NeRV models in Sec.~\ref{4_3}, discuss the significance of shortcuts and the multi-scale structure in the proposed PNeRV in Sec.~\ref{4_4}.

\vspace{-0.02in}
\subsection{Basic Definitions and Notations}
\vspace{-0.02in}
One of the main issues for the UAT analysis of a finite length $L$ feed-forward network is to find out the minimal width $w^*:=\min \max d_i, 1 \leq i \leq L$ where $d_i$ is the width of the $i$-th layer so that neural nets with width $w^*$ and length $L$ can approximate any scalar continuous function arbitrarily well~\cite{ApproximatingCF, UniversalFA, MinimumWF}. Following the statement in~\cite{ApproximatingCF}, a deep affine net is defined as follows.
\vspace{-0.02in}
\begin{definition}(Deep Affine Net).\label{N}
\vspace{-0.02in}
A deep affine net of $L$ layers is given as follows:
\begin{equation}\small
\mathcal{N} := A_L \circ \sigma \circ A_{L-1} \circ \cdots \circ \sigma \circ A_{1}.
\end{equation}
where the $i^{th}$ layer is an affine transformations $A_{i} := \mathbb{R}^{d_i} \to \mathbb{R}^{d_{i+1}}$, $d_1 = d_{in}$, $d_L = d_{out}$ with $\sigma$ as activation. 
\end{definition}\vspace{-0.03in}
In existing NeRV research,
NeRV~\cite{nerv} and HNeRV~\cite{HNeRV} meet this definition.

\subsection{Implicit Neural Video Coding}\label{invc_sec}

Recently, INR-based video coding has received increasing attention, and it uses a lightweight model to fit a video clip. We formulate this coding pipeline as \textit{Implicit Neural Video Coding} (INVC), and the decoder with its embeddings together is known as the \textit{NeRV system}~\cite{nerv, enerv, HNeRV, divnerv, DNeRV}. 
\begin{definition}(NeRV System).\label{D2}
~Each frame ${V_t}$ in an RGB video clip $V=\{ {V_t} \}^T_{t=1} \in \mathbb{R}^{T\times 3 \times H \times W}$ is represented by an implicit unknown continuous function $\mathcal{F}: [0,1]^{d_{in}} \to \mathbb{R}^{d_{out}}$ with the embedding $\mathcal{E}(t)$ obtained by encoder $\mathcal{E}: \mathbb{N} \to [0,1]^{d_{in}} $ on the $t$ time stamp,
\[\small
V_t =  \mathcal{F} \circ \mathcal{E} \left( t \right),
\]
where $\mathcal{F}$ can be approximated by a learnable neural network $\mathcal{D}$ of finite length $L_{\mathcal{D}}$, width $w_{\mathcal{D}}$ and activation $\sigma$. The reconstruction $\tilde{V_t}$ via $\mathcal{D}$ and $\mathcal{E}$ is given as follows:
\begin{gather}\label{recons}\small
\tilde{V_t} = \mathcal{D} \circ \mathcal{E} \left( t \right),  \nonumber 
\end{gather}
where the decoder $\mathcal{D}$ and embedding $\mathcal{E}(t)$ together are known as NeRV system, $\{ \mathcal{D}, \mathcal{E} (t) \}^{T}_{t=1}$. 
\end{definition}\vspace{-0.05in}

For the index-based models~\cite{nerv} and~\cite{enerv}, the encoder $\mathcal{E}$ is Positional Encoding~\cite{siren}. In content-based models~\cite{HNeRV, DNeRV}, $\mathcal{E}$ is learnable and provides content embeddings. When $\mathcal{D}$ is a deep affine net, it is named as a \textit{serial cascaded NeRV system}, such as NeRV~\cite{nerv} and HNeRV~\cite{HNeRV}, and $\mathcal{D}$ is formulated as follows, where $B_{l}$ is the $l$-th upsampling layer. 
\begin{gather}\label{recons}\small
\mathcal{D} := B_L \circ \sigma \circ B_{L-1} \circ \cdots \circ \sigma \circ B_{1}.
\end{gather}
We present the proposed \textbf{I}mplicit \textbf{N}eural \textbf{V}ideo \textbf{C}oding \textbf{P}roblem (\textbf{INVCP}) as follows. More discussions between INVCP and existing pipelines are given in Sec.~{\color{red}A}.
\vspace{-0.05in}
\begin{problem}(INVCP).\label{IVNCP}
The goal of INVC is to obtain the minimal parameter quantity under a certain approximation error $\epsilon$ between input $V$ and reconstruction $\tilde{V}$,
\begin{small}
\begin{gather}\label{IVNC}
\mathop{\arg\min}_{\mathcal{D}, \mathcal{E} }~ \mathsf{Param} \left( \mathcal{D} \right) + \sum_{t=1}^{T} d_{in}^t , \nonumber  \\
\mathrm{ s.t. } ~ L_{\mathcal{D}}, w_{\mathcal{D}} \in \left[1, \infty \right),~\mathop{\sup} \mathop{\sum} \Vert \tilde{V_t} - V_t\Vert \leq \epsilon,~t \in [1, T]. \notag
\end{gather}
\end{small}where $d_{in}^t$ is the dimension of embedding $\mathcal{E}(t)$ w.r.t. the $t$-th frame, $L_{\mathcal{D}}$ and $w_{\mathcal{D}}$ are the length and width. 
\end{problem}\vspace{-0.05in}


\vspace{-0.02in}
\subsection{UAT Analysis of Cascaded NeRV Model}\label{4_3}
\vspace{-0.02in}

For video INRs, the model strives to capture the implicit function that efficiently encodes a video. Within the UAT framework, a keen focus is on the smoothness properties of this implicit function, as it also encapsulates the video's inherent dynamics. 

We name these properties as \textit{rate of dynamics}, referring to the differences and transitions between consecutive frames within the video. We introduce $\omega_{\mathcal{V}}^{-1}$ to informally represent the rate of dynamics for video $\mathcal{V}$, inspired by the mathematical techniques used in UAT analysis~\cite{ApproximatingCF}.

\vspace{-0.05in}
\begin{definition}\label{moc}
The dual modulus of continuity $\omega_f^{-1}$ w.r.t. a continuous $f$ defined on $\Omega$ is set as
\[\small
\omega_f^{-1}(\epsilon)  := \mathop{\sup} \{ \delta : \omega_f (\delta) \leq \epsilon \},
\]
where $\omega_f$ represents the modulus of continuity of $f$
\[\small
\omega_f(\delta) := \mathop{\sup}_{x,y \in\Omega} \{ \Vert f(x)-f(y)\Vert : d(x,y) \leq \delta \}.
\]
\end{definition}\vspace{-0.05in}
\begin{proposition}\label{r1}
Using a function $\mathcal{F}: \mathbb{N} \to \mathbb{R}^{d_{V}}$ to roughly represent a video $V$, when the variation of frames (video dynamics) $\Vert \mathcal{F}(t_i)-\mathcal{F}(t_j) \Vert$ is at a certain level $\epsilon$ for two time stamps $t_i$ and $t_j$, then the longer the duration sustains, the larger $\omega_\mathcal{F}^{-1}$ gets. Smoother video has larger $\omega_\mathcal{F}^{-1}$.
\end{proposition}\vspace{-0.05in}

Notably, the explicit calculation of $\omega_f^{-1}$ is hard to obtain, and it is more like an empirical judgment, such as camera movement, subject speed, noise, and others. We present the estimation of the upper bound of the minimal parameter quantity of the cascaded NeRV model as Theorem~\ref{param}. The proof of Theorem~\ref{param} can be found in Sec.~{\color{red}A.3}.

\vspace{-0.03in}
\begin{theorem}\label{param}
For a cascaded NeRV system to $\epsilon$-approximate a video $V$ which is implicitly characterized by a certain unknown L-Lipschitz continuous function $\mathcal{F}: K \to \mathbb{R}^{d_{out}}$ where $K \subseteq \mathbb{R}^{d_{in}}$ is a compact set, then the upper bound of the minimal parameter quantity $\mathsf{Param}(\mathcal{D})$ is given as\vspace{-0.02in}
\[\small
 \mathsf{Param}_{\min}(\mathcal{D})\leq d_{out}^2 \left( \frac{ \mathop{O}
    \left(diam  \left(K \right)
    \right) }{\omega_\mathcal{F}^{-1} \left( \epsilon \right) }
\right)^{d_{in}+1}.
\]
\end{theorem}

From Theorem~\ref{param}, it can be seen that for a video, the fitting performance of the cascaded NeRV model depends on the rate of dynamics $\omega_\mathcal{F}^{-1}$ and the dimension of the video, $d_{out}$. The smoother and lower the dimension of the video to be modeled, the less difficult it is to approximate.

\vspace{-0.03in}
\begin{proposition}\label{AtoP}
The rate of dynamics for a given video will determine the performance of the NeRV system.
\end{proposition}

\vspace{-0.03in}
\subsection{UAT Analysis of PNeRV}\label{4_4}
\vspace{-0.03in}

According to Theorem~\ref{param}, the upper bound of parameters of cascaded NeRV required for model fitting only depends on the properties of the target video. It demonstrates that, although different models can exhibit diverse architectures, their fitting behavior on the same video tends to be similar, indicating a limitation in the model's ability. However, according to observations in UAT research~\cite{ResNetWO, UniversalAB}, the model with shortcuts will reduce the maximum width to 1, indicating that the model size can be greatly reduced while maintaining the performance. Therefore, the involvement of \textbf{shortcuts} is the key to enhancing model capability. 

Besides, we believe the implicit function representing a video can be decomposed into diverse sub-functions from a pattern-disentangled perspective. If we treat each stream in $\mathcal{D}$ as a sub net, the whole $\mathcal{D}$ is an ensemble,
\begin{equation}\label{sub_net}\small
\mathcal{D} := \sum A_L^{(i)} \circ \rho_{L-1}^{(i)} \circ A_{L-1}^{(i)} \circ \cdots \circ \rho_{1}^{(i)} \circ A_{1}^{(i)}.
\end{equation}

Different shortcut pathways can fit various patterns, as a single shortcut has the universal approximation ability. For example, in Fig.~\ref{pnerv}, \ding{172} may capture the low-frequency motions. Whereas \ding{173}, directed towards fine-grained layers, signifies spatial details. 
This hypothesis aligns with the empirical evidence observed in other vision areas, which shows that the pyramid structure, a widely adopted hierarchical topology, can improve dense prediction tasks. 
That is why PNeRV outperforms others and achieves less semantic uncertainty and better perceptual quality. 
\begin{proposition}\label{pro2}\vspace{-0.03in}
As the ensemble of sub-nets, the Pyramidal structure will enhance the perceptual quality of NeRV systems. \vspace{-0.05in}
\end{proposition}
\vspace{-0.02in}
\section{Experiment}
\vspace{-0.02in}

\textbf{Settings}. We perform video regression on 2 datasets, and all videos are center cropped to a $1 \times 2$ ratio. UVG~\cite{uvg} has 7 videos 
with a size of $960 \times 1920$ in $300$ or $600$ frames at 120 FPS. DAVIS~\cite{davis} is a large dataset of 47 videos in $960 \times 1920$, containing large motions and complex spatial details. 
We choose 9 videos\footnote{Bmx-bumps, Camel, Dance-jump, Dog, Drift-chicane, Elephant, Parkour, Scooter-gray, Soapbox.} from DAVIS as a subset, containing different types of spatiotemporal features. 

\noindent\textbf{Metrics}. We use PSNR and MS-SSIM to evaluate pixel-wise errors. For spatial consistency, we choose the Learned Perceptual Image Patch Similarity (LPIPS)~\cite{LPIPS} and Frechet Video Distance (FVD)~\cite{FVD} as perceptual metrics, where LPIPS is based on AlexNet and FVD is based on the I3D model. The difference between PNeRV (P) and the baseline (B) is calculated as $(B-P)/B$ to show the improvement.

\noindent\textbf{Training}. We adopt Adam as the optimizer, where beta is (0.9, 0.999) and weight decay is 0. The learning rate is 5e-4 with a cosine annealing schedule. The loss function is L2, and the batch size is 1. All experiments are conducted using PyTorch 1.8.1 on NVIDIA GPU RTX2080ti, training for 300 epochs. We choose NeRV~\cite{nerv}, E-NeRV~\cite{enerv}, HNeRV~\cite{HNeRV}, DivNeRV~\cite{divnerv} and DiffNeRV~\cite{DNeRV} as baseline models. All models are trained with a similar 3M size, and we follow the setting of embedding size as the baseline method.

\vspace{-0.02in}
\subsection{Video Regression on UVG}
\vspace{-0.02in}
\textbf{Pixel-wise error}. PSNR comparison on UVG is reported in Tab.~\ref{uvg_table}, where bold font is the best result and underline is the second best. PNeRV-L surpasses other models (+0.42 dB against DiffNeRV and +4.25 dB against NeRV).
PNeRV-M achieves the best result against other single-stream models (+1.96 dB against HNeRV and +3.02 dB against NeRV). 
The proposed pyramidal architecture shows its effectiveness when combined with various encoders.

\noindent\textbf{Perceptual quality}. The perceptual results are given in Tab.~\ref{lpips} (LPIPS) and Tab.~\ref{uvg_fvd} (FVD), and the results of PNeRV show a significant improvement, especially for ``Bospho'' and ``ShakeN''. The FVD results in Tab.~\ref{uvg_fvd} indicate that PNeRV provides better spatiotemporal consistency compared to other baseline models (+231\% against NeRV~\cite{nerv} and +64.5\% against DiffNeRV~\cite{DNeRV}). 

\noindent\textbf{Case study}. The visualized comparison on UVG is exhibited in the bottom three rows of Fig.~\ref{davis_detail}. For dynamic objects with indistinct boundaries or noisy backgrounds, such as the horse in ``ReadyS" and the tail in ``ShakeN," PNeRV demonstrates superior visual quality without requiring additional semantic information.

\begin{table*}[!ht]
      \centering
    \setlength{\abovecaptionskip}{2pt}
    \setlength{\belowcaptionskip}{-5pt}
       \tabcolsep=0.25cm
        \resizebox{!}{2.1cm}{
        \begin{tabular}{l||c|c|ccccccc|c}
      \textit{PSNR} $\uparrow$  & \textsf{D.P.} & \textsf{E.S.} & Beauty & Bospho & HoneyB & Jockey & ReadyS & ShakeN &YachtR &Avg.~M.\\
    \midrule[1.5pt]
    Avg.~V.  & N/A & N/A  &36.06 & 35.32 & \textcolor{blue}{39.48} &33.27 &\textcolor{red}{27.53} &35.27 &30.03 & N/A \\
    \midrule
    NeRV~\cite{nerv}  &3M   &160  & 33.25   & 33.22   & 37.26  &31.74  & 24.84 & 33.08  & 28.03 &31.63  \\
     $\text{NeRV}^*$~\cite{nerv}  &3.2M &160  & 32.71   & 33.36   & 36.74  &32.16  & 26.93 & 32.69  & 28.48 &31.87  \\
    E-NeRV~\cite{enerv}    &3M  &160  &33.17    & 33.69   & 37.63  &31.63  & 25.24 &34.39 	 & 28.42 &32.02 \\
    HNeRV~\cite{HNeRV}     &3M  &128  &33.58    & 34.73   & 38.96  &32.04  & 25.74 &34.57   & 29.26 & 32.69\\
    DiffNeRV~\cite{DNeRV}   &3.4M &6528 &\textbf{40.00} &36.67 &41.92 &\underline{35.75} &28.67  &36.53   &\underline{31.10} &\underline{35.80}\\
   $\text{DivNeRV}^*$~\cite{divnerv} &3.2M &N/A &33.77    &\textbf{38.66}   &37.97   &35.51   &\textbf{33.93}  &35.04   &\textbf{33.73}  &35.52\\
    \midrule[1pt]
     PNeRV-M &1.5M &128 &37.51 & 33.80  &41.76  &29.96    & 24.15  &36.18 & 28.92 & 33.18\\
     &3M &128 &39.08 & 35.56  &\underline{42.59} &31.51 & 25.94  &\underline{37.61} & 30.27 &34.65\\
 \midrule
     PNeRV-L &1.5M &6528 &37.98  & 35.18  &41.78   &34.43 & 27.28    &36.65  & 28.29 & 34.51\\
    &3.3M &6528 &\underline{39.46}  & \underline{36.68}  &\textbf{42.73}  &\textbf{35.81} & \underline{28.97}   &\textbf{38.25} & 30.92 & \textbf{36.12}  
    \\
         \end{tabular}
        }
    \caption{PSNR comparison on UVG: the larger, the better. $*$ indicates methods that fit videos in a shared model while others fit each video in a single model. \textsf{D.P.} is the parameter quantity of the decoder, and \textsf{E.S.} is the corresponding embedding size per frame. Avg.~V is the average PSNR across all models for the same video. Avg.~M is the average PSNR for a single model on the entire dataset. 
    }\label{uvg_table}
\end{table*}

\begin{table*}[!t]\scriptsize
     \centering
    \setlength{\abovecaptionskip}{2pt}
    \setlength{\belowcaptionskip}{-5pt}
       \tabcolsep=0.1cm
  \resizebox{!}{1.15cm}{\centering
  \begin{tabular}{l||l|l|l|l|l|l|l|l|l|c}
    \textit{PSNR} / \textit{SSIM} $\uparrow$    &Bmx-B & Camel & Dance-J & Dog & Drift-C &Elephant &Parkour &Scoo-gray  &Soapbox  &Avg.\\
    \midrule[1.2pt]
    NeRV~\cite{nerv}       &29.42/0.864  &24.81/0.781  &27.33/0.794  &28.17/0.795 &36.12/0.969 &26.51/0.826	 &25.15/0.794 &28.16/0.892 &27.68/0.848 &27.99/0.840 \\
    E-NeRV~\cite{enerv}   &28.90/0.851  & 25.85/0.844 & 29.52/0.855 &30.40/0.882 &39.26/0.983 &28.11/0.871  &25.31/0.845  &29.49/0.907 &28.98/0.867 &29.62/0.878 \\
    HNeRV~\cite{HNeRV}  &29.98/0.872 & 25.94/0.851 &29.60/0.850 &30.96/0.898 &39.27/0.985 &28.25/0.876 &26.56/0.851 &31.64/0.939  &29.81/0.881 &30.22/0.889 \\
    DiffNeRV~\cite{DNeRV} &30.58/0.890  & 27.38/0.887 &29.09/0.837 &\textbf{31.32/0.905} &\textbf{40.29}/0.987 &27.30/0.848 & 25.75/0.827 &30.35/0.923 &\textbf{31.47/0.912} &30.39/0.890 \\
 \midrule
    PNeRV-L (ours)   &\textbf{31.05/0.896}  & \textbf{27.89/0.892} & \textbf{30.45/0.873} &31.08/0.898 &40.23/\textbf{0.987} &\textbf{29.72/0.903} &\textbf{27.53/0.878} &\textbf{32.68/0.950} &30.85/0.902 & \textbf{31.27/0.908} \\
  \end{tabular}}
     \caption{PSNR and MS-SSIM comparison on DAVIS.}\label{davis}
\end{table*}

\noindent\textbf{Compared with the SOTA}. As shown in Tab.~\ref{uvg_table}, PNeRV obtained competitive PSNR results on dynamic and smooth videos. 
\cite{divnerv} is less effective for videos with fewer motions but complicated contextual spatial correlation.
Also, \cite{DNeRV} makes it hard to reconstruct the videos filled with high-frequency details.
By comparison, PNeRV achieves comparable performance on all videos.

\vspace{-0.02in}
\subsection{Video Regression on DAVIS}
\vspace{-0.02in}
\noindent\textbf{Pixel-wise error}. In Tab.~\ref{davis}, we present the PSNR and SSIM comparison on the DAVIS dataset. 
PNeRV gains a +0.88 dB PSNR increase compared to DiffNeRV and +3.28 dB compared to vanilla NeRV. Despite the challenges posed by complex spatiotemporal features, PNeRV exhibits significant improvements (refer to ``Parkour'', which is the most difficult one, or ``Drift-chicane'', where the racing car undergoes intense motion amidst smoke-induced noise).

\noindent\textbf{Perceptual quality}. The LPIPS results on DAVIS are reported in Tab.~\ref{lpips}, where PNeRV achieved a 32.0\% increase compared to NeRV and 12.6\% against the second-best DiffNeRV. In Tab.~\ref{davis_fvd}, PNeRV gains a 634\% FVD increase over NeRV and 128\% against DiffNeRV. For the worst case, ``Dog'', although PNeRV obtained a poor FVD result owing to the severe global blurring caused by camera motion, the PSNR is only slightly lower than the best (-0.24 db).

\begin{figure*}[t]
  \centering
    \setlength{\abovecaptionskip}{2pt}
    \setlength{\belowcaptionskip}{-5pt}
  \includegraphics[width=0.88\linewidth]{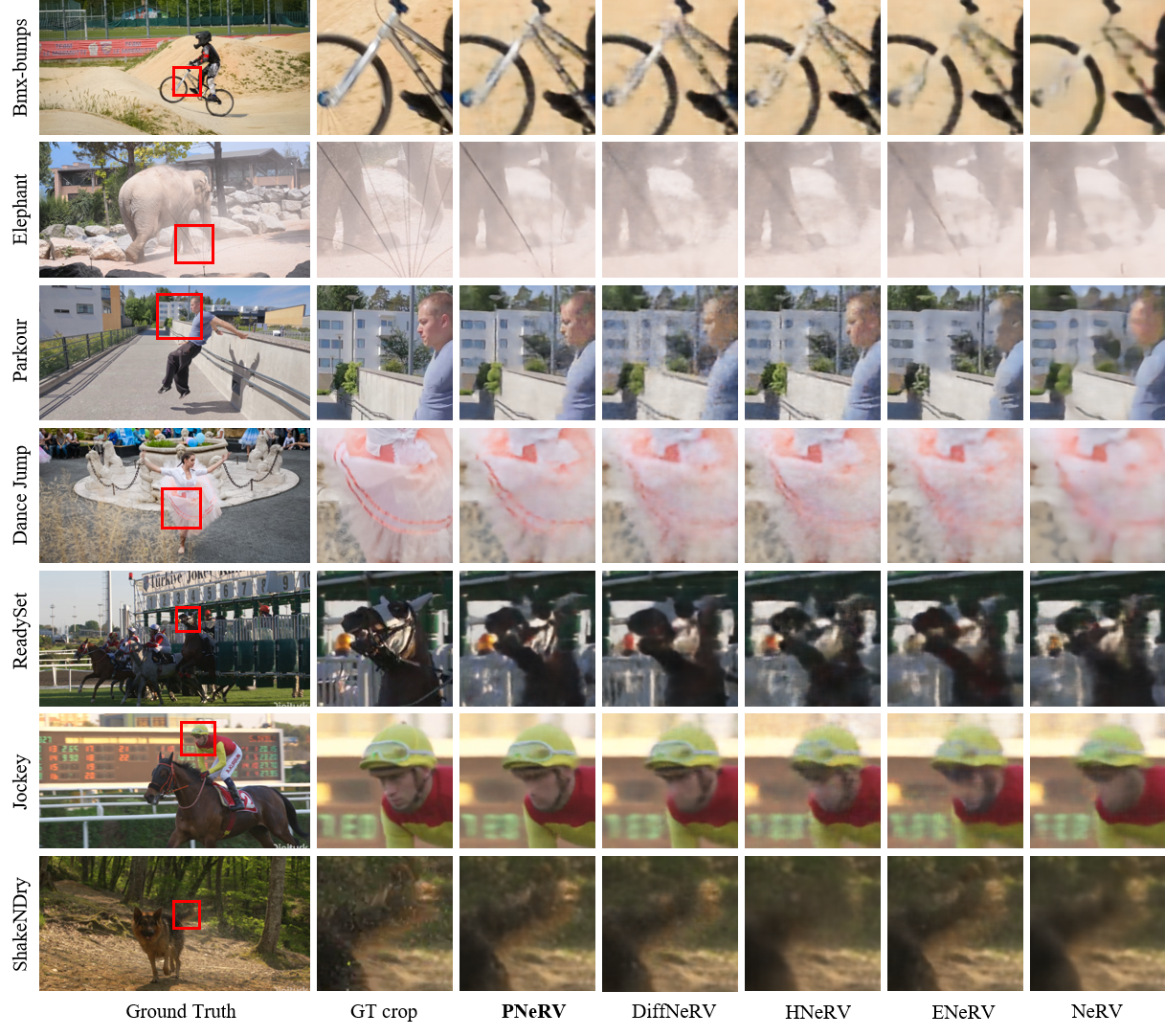}\\
  \vspace{-0.05in}\caption{Visual comparison on various videos. ``Bmx'' has larger motion, ``Elephant'' has massive droplets blurring, ``Parkour'' involves both camera rotation and extreme dynamics, ``Dance'' contains large motion under high-frequency reed leaves. ``Jockey'', ``ReadyS'', and ``ShakeN'' are videos with complex spatiotemporal correlation in UVG. Zoom in for a detailed comparison. 
  }
  \label{davis_detail}
\end{figure*}

\noindent\textbf{Case study}. Visualizations are shown in Fig.~\ref{davis_detail}. PNeRV reduced spatial inconsistency, particularly in ``Dance Jump" and ``Elephant," which are filled with irrelevant high-frequency details obscuring semantic clarity.

\begin{table*}[!h]
    \begin{subtable}{.5\linewidth}
      \centering
       \tabcolsep=0.1cm
        \resizebox{!}{1.0cm}{
  \begin{tabular}{l||ccccccc|c}
 \textit{LPIPS}    $\downarrow$      & Beauty & Bospho & HoneyB & Jockey & ReadyS & ShakeN &YachtR &Avg.\\
 \midrule[1.5pt]
    NeRV~\cite{nerv}      &0.229 &0.203 &0.043 &0.251 &0.326 &0.189 &0.276 &0.216\\
    ENeRV~\cite{enerv}    &0.224 &0.179 &0.039 &0.279 &0.318 &0.168 &0.363 &0.224\\
    HNeRV~\cite{HNeRV}    &0.218 &0.172 &0.042 &0.270 &0.348 &0.191 &0.253 &0.213\\
    DiffNeRV~\cite{DNeRV} &\textbf{0.205} &0.164 &0.042 &0.196 &\textbf{0.206} &0.181 &0.241 &0.176\\
  \midrule[0.1pt]
    PNeRV (ours)          &0.210 &\textbf{0.132} &\textbf{0.037} &\textbf{0.177} &0.211 &\textbf{0.146} &\textbf{0.230} &\textbf{0.163}\\
     \end{tabular}
        }
        \label{uvg_lpips}
    \end{subtable}%
      \begin{subtable}{.5\linewidth}
      \centering
       \tabcolsep=0.1cm
        \resizebox{!}{1.0cm}{
   \begin{tabular}{l||ccccccccc|c}
 \textit{LPIPS}    $\downarrow$      & Bmx-B & Camel & Dance & Dog & Drift &Eleph & Parko &Scoo-g &Soapb &Avg.\\
 \midrule[1.5pt]
    NeRV~\cite{nerv}      &0.374   &0.476           &0.517          &0.573          &0.136    &0.490      &0.481 &0.308  &0.424 &0.419\\
    ENeRV~\cite{enerv}    &0.386   &0.357           &0.426          &0.404          &0.061    &0.419      &0.429 &0.282  &0.380 &0.349\\
    HNeRV~\cite{HNeRV}    &0.315   &0.331           &0.392          &0.405          &0.058    &0.387     &0.414  &0.226 &0.357 &0.321 \\
    DiffNeRV~\cite{DNeRV} &0.320   &\textbf{0.278}  &0.423          &0.394   &\textbf{0.053} &0.431 &0.478 &0.268  &\textbf{0.297} &0.326\\
  \midrule[0.1pt]
    PNeRV (ours)   &\textbf{0.308} &0.284    &\textbf{0.363} &\textbf{0.387}     &0.054   &\textbf{0.343}  &\textbf{0.314} &\textbf{0.188} &0.324 &\textbf{0.285}\\
  \end{tabular}
    }
    \label{davis_lpips}
    \end{subtable}
   \vspace{-0.1in} 
   \caption{LPIPS comparison on UVG (left) and DAVIS (right) dataset. }\label{lpips}
\end{table*}

\begin{table*}[!h]\small
   \tabcolsep=0.2cm
    \setlength{\abovecaptionskip}{2pt}
    \setlength{\belowcaptionskip}{-5pt}
      \centering
      \resizebox{!}{1.1cm}{
       \begin{tabular}{l||lc|lc|lc|lc|lc|lc|lc|c}
        \textit{FVD}$\downarrow$ \ \textit{Gap}$\uparrow$ & \multicolumn{2}{c|}{Beauty} & \multicolumn{2}{c|}{Bospho} & \multicolumn{2}{c|}{HoneyB} & \multicolumn{2}{c|}{Jockey} & \multicolumn{2}{c|}{ReadyS} & \multicolumn{2}{c|}{ShakeN} & \multicolumn{2}{c|}{YachtR} & \multicolumn{1}{c}{Avg. $\uparrow$}\\

    \midrule[1.5pt]
    NeRV~\cite{nerv}    &3.76e-5 &281\%  &1.00e-4&253\%  &1.45e-5&193\%  &5.81e-4&499\%  &1.98e-3&122\%   &3.27e-5&178\%  &4.07e-4&92.8\% &231\%\\
    ENeRV~\cite{enerv}  &2.66e-5 &169\%  &7.86e-5&176\%  &5.88e-6&186\%  &1.00e-3&936\%  &1.46e-3&64.2\%  &2.12e-5&80.7\% &1.00e-3&376\%  &284\%\\
    HNeRV~\cite{HNeRV}  &3.29e-5 &233\%  &6.74e-5&137\%  &1.50e-5&203\%  &9.46e-4&874\%  &2.07e-3&132\%   &5.06e-5&331\%  &3.56e-4&68.8\% &282\%\\
    DiffNeRV~\cite{DNeRV}&1.29e-5 &30.7\% &4.28e-5&50.3\% &6.50e-6&31.1\% &1.55e-4&60.1\% &\textbf{6.58e-4}&-26.3\% &4.69e-5&300\%  &2.23e-4&5.9\%  &64.5\%\\
      \midrule[0.1pt]
    PNeRV     (ours)    &\textbf{9.88e-6} & -     &\textbf{2.85e-5}&-      &\textbf{4.96e-6}&-    &\textbf{9.70e-5}&-    &8.94e-4&-     &\textbf{1.17e-5}&-    &\textbf{2.11e-4}&-    &-\\
  \end{tabular}}
  \caption{FVD comparison on UVG.}\label{uvg_fvd}
\end{table*}

\begin{table*}[!h]\small
   \tabcolsep=0.08cm
    \setlength{\abovecaptionskip}{2pt}
    \setlength{\belowcaptionskip}{-5pt}     \centering
    \resizebox{!}{1.05cm}{
  \begin{tabular}{l||lc|lc|lc|lc|lc|lc|lc|lc|lc|c}
     \textit{FVD}$\downarrow$ \ \textit{Gap}$\uparrow$ & \multicolumn{2}{c|}{Bmx-B } & \multicolumn{2}{c|}{Camel} & \multicolumn{2}{c|}{Dance-Jump} & \multicolumn{2}{c|}{Dog} & \multicolumn{2}{c|}{Drift-C} & \multicolumn{2}{c|}{Elephant} & \multicolumn{2}{c|}{Parkour} & \multicolumn{2}{c|}{Scoo-gray} & \multicolumn{2}{c|}{Soapbox} & \multicolumn{1}{c}{Avg. $\uparrow$}\\
    \midrule[1.5pt]
    NeRV~\cite{nerv}     &8.99e-5 &146\%  &2.70e-4&404\%    &6.66e-5&1273\%  &3.02e-5&336\%  &3.85e-6&2830\%  &2.470e-5 &95.8\%  &1.35e-4&309\%  &3.815e-5 &197\%     &9.39e-5&115\% &634\%\\
    ENeRV~\cite{enerv}  &1.20e-4 &229\%  &1.08e-4&102\%    &6.05e-6&24.8\%  &4.04e-6 &-41.5\%  &5.41e-7&311\%  &2.647e-5 &110\%  &7.09e-5&114\%  &3.961e-5 &208\%  &7.01e-5&61.1\%  &124\%\\
    HNeRV~\cite{HNeRV}  &4.97e-5 &36.2\%  &1.04e-4&94.1\%  &9.58e-6&97.5\%  &4.51e-6&-34.6\%  &1.21e-6&821\%  &4.439e-5 &252\% &7.81e-5&135\%  &2.256e-5 &75.8\%    &7.36e-5&69.3\% &171\%\\
    DiffNeRV~\cite{DNeRV}&\textbf{3.11e-5} &-14.8\% &\textbf{3.85e-5}&-28.1\% &1.19e-5&146\%  &\textbf{3.61e-6}&-47.6\% &6.48e-7&392\%  &6.408e-5 &408\%   &1.45e-4&339\%  &1.614e-5 &25.7\%  &\textbf{1.64e-5}&-62.2\%  &128\%\\
   \midrule[0.1pt]
    PNeRV     (ours)     &3.65e-5 & -     &5.36e-5&-        &\textbf{4.85e-6}&-      &6.91e-6&-    &\textbf{1.31e-7}&-  &\textbf{1.261e-5}&-   &\textbf{3.31e-5}&-  &\textbf{1.283e-5}&-    &4.35e-5&-    &-\\
  \end{tabular}
  }
  \caption{FVD comparison on DAVIS.}\label{davis_fvd}
\end{table*}

\vspace{-0.02in}
\subsection{Ablation Studies}
\vspace{-0.02in}

The ablation of the effectiveness of the proposed pyramidal architecture is in Tab.~\ref{abla_ms}, and the contributions of two proposed modules are validated in Tab.~\ref{abla_da}, where the parameters of different models remain the same for a fair comparison.

\noindent\textbf{Overall structure}. We validate the design of the multi-level structure on the most dynamic and smooth videos (``Parkour'' and ``HoneyB''). In Tab.~\ref{abla_ms}, the ``serial'' in the first row represents HNeRV~\cite{HNeRV}.
``Pyram.+Concat." incorporates solely shortcuts without fusion modules. The main difference between DiffNeRV and PNeRV-L is the quantity of shortcuts (2 vs 5), and PNeRV-L performs better. 

\noindent\textbf{Modules contribution}. We compare \textsc{KFc} with two upscaling layers, Deconv~\cite{deconv} and Bilinear (the combination of bilinear upsampling and Conv2D). \textsc{KFc} performs better due to the global receptive field, as shown in Tab.~\ref{abla_da}.

Also, we compare BSM with Concat, GRU~\cite{gru} and LSTM~\cite{LSTM}. The results suggest that, disentangled feature fusion significantly enhances performance. Detailed results for each video are listed in Tab.~{\color{red}C.6} in the appendix.

\begin{table*}[t]\scriptsize
      \centering
    \setlength{\abovecaptionskip}{2pt}
    \setlength{\belowcaptionskip}{-5pt}    \tabcolsep=0.25cm
   \resizebox{!}{1.35cm}{\centering
 \begin{tabular}{l||lll|l||lll|l}
  \multirow{2}*{\shortstack{~}} &\multicolumn{4}{c||}{Parkour (\textit{Dynamic})}  &\multicolumn{4}{c}{HoneyB (\textit{Smooth})}  \\
     Models Size             &1.5M & 3M & 5M & Avg. & 0.75M & 1.5M & 3M & Avg.\\
    \midrule[1.2pt]
    Serial    (HNeRV~\cite{HNeRV})           &25.07  & 26.56  &24.34 &25.32  &36.65 & 36.72 & 38.96  &37.44\\
    Pyram. + Concat.                         &24.20  & 25.45  &25.83 &25.16  &40.07 & 41.58 & 42.34  &41.33\\
    Pyram. + BSM.    (\textbf{PNeRV-M})      &24.81  & 26.02  &27.13 &25.99  &40.34 & 41.36 & 42.59  &41.43\\
    \midrule[0.5pt]
    Serial + Diff.   (DiffNeRV~\cite{DNeRV}) &25.49  & 25.75  &25.71 &25.65  &\textbf{40.52}& 41.52  &41.92 &41.32\\
    Pyram. + Diff. + BSM. (\textbf{PNeRV-L}) &\textbf{25.62}  & \textbf{27.08}  &\textbf{27.21}  &\textbf{26.67} &39.81 & \textbf{41.85} &\textbf{42.73} &\textbf{41.46}\\
  \end{tabular}
  }
  \caption{Ablation studies for model size and overall architecture on ``HoneyB'' and ``Parkour''.}\label{abla_ms}
\end{table*}

\vspace{-0.02in}
\subsection{Validation of Theoretical Analysis}
\vspace{-0.02in}

The results in Tab.~\ref{uvg_table} and Tab.~\ref{abla_ms} validate the Remark~\ref{AtoP}. For those smooth videos with larger $\omega_f^{-1}$ and a smaller upper bound, models may obtain better performance; vice versa. 
The results of PNeRV in Fig.~\ref{davis_detail}, which exhibit less noise and blurring, validate Remark.~\ref{pro2}. 
Hierarchy structure reduces ambiguity and artifacts caused by semantic uncertainty.

\begin{table*}[!h]\scriptsize
\setlength\belowdisplayskip{-0.02in}
      \centering
    \setlength{\abovecaptionskip}{2pt}
    \setlength{\belowcaptionskip}{-5pt} 
       \tabcolsep=0.2cm
  \resizebox{!}{0.9cm}{\centering
 \begin{tabular}{l||llll}
   PSNR$\uparrow$ SSIM$\uparrow$ (\textcolor{red}{A.P.G.})$\uparrow$   &Concat &GRU  & LSTM & \textbf{BSM}\\
    \midrule[1.2pt]
    Bilinear       &27.16/0.816(\textcolor{red}{-4.14})&28.39/0.847(\textcolor{red}{-2.91})  &28.07/0.834(\textcolor{red}{-3.23}) &29.08/0.862(\textcolor{red}{-2.22})  \\
    Deconv         &27.37/0.803(\textcolor{red}{-3.93})&29.00/0.845(\textcolor{red}{-2.30})  &28.91/0.850(\textcolor{red}{-2.39}) &29.96/0.881(\textcolor{red}{-1.34})  \\
    \textbf{KFc}   &28.68/0.848(\textcolor{red}{-2.62})&29.31/0.868(\textcolor{red}{-1.99})  &29.04/0.866(\textcolor{red}{-2.26}) &\textbf{31.30/0.904}(\textcolor{blue}{+0})  \\
  \end{tabular}}
   \caption{Contribution ablations for \textsc{KFc} and BSM, reported as average results on 7 DAVIS videos. A.P.G. indicates the average PNSR gap compared with the final version of PNeRV (\textsc{KFc} + BSM); the larger the better. Detailed results for each video are given in Sec.~{\color{red}D.2}.}\label{abla_da}
\end{table*}

\vspace{-0.02in}
\subsection{Additional Experiment Results}
\vspace{-0.02in}
Additional results are provided in the appendix.
Video interpolation on UVG is discussed in Sec.~{\color{red}C.1} where PNeRV achieves the second-best PSNR (31.18 dB), exceeding the vanilla NeRV (26.54 dB). Video compression is shown in Sec.~{\color{red}C.2}, where competitive results are achieved over different coding pipelines. 
Video inpainting on the DAVIS subset is provided in Sec.~{\color{red}C.3}, where an average PSNR of 25.54 dB is achieved, outperforming NeRV (22.71 dB) and DNeRV (25.20 dB).
More visual examples are shown in Sec.~{\color{red}C.4}, and visualization of feature maps in Sec.~{\color{red}D.1}. More detailed ablations are presented in Sec.~{\color{red}D.2}. More video examples with the link are listed in Sec.~{\color{red}C.6}.

\vspace{-0.05in}
\section{Conclusion}
\vspace{-0.05in}
To resolve the spatiotemporal inconsistency issue, we propose Pyramidal NeRV realizing multi-level information interaction by a low-cost \textsc{KFc} and a fusion module BSM. Further, we use UAT to provide some explanations and insights for NeRV. Competitive results on various tasks and metrics validate the superiority of PNeRV.

\noindent\textbf{Limitation and future work}. Hierarchical structure brings higher computational complexity. We will optimize redundant modules of the model for acceleration in the future.

\noindent\textbf{Acknowledgements}. The work was supported in part by
National Key Research and Development Project of China
(2022YFF0902402) and U.S. National Science Foundation award (CCF-2046293). 
{
    \small
    \bibliographystyle{ieeenat_fullname}
    \bibliography{main}
}
\clearpage
\setcounter{page}{1}
\setcounter{section}{0}
\setcounter{table}{0}
\setcounter{figure}{0}
\setcounter{theorem}{0}
\setcounter{definition}{0}
\setcounter{lemma}{0}
\setcounter{proposition}{0}
\setcounter{remark}{0}
\setcounter{problem}{0}

\renewcommand\thesection{\Alph{section}}
\maketitlesupplementary
\renewcommand\thefigure{\thesection{}.\arabic{figure}}
\renewcommand\thetable{\thesection{}.\arabic{table}}
\renewcommand\thetheorem{\thesection{}.\arabic{theorem}}
\renewcommand\thedefinition{\thesection{}.\arabic{definition}}
\renewcommand\theproposition{\thesection{}.\arabic{proposition}}
\renewcommand\thelemma{\thesection{}.\arabic{lemma}}
\renewcommand\theremark{\thesection{}.\arabic{remark}}
\renewcommand\theproblem{\thesection{}.\arabic{problem}}


\section{More Discussions of Universal Approximation Theory (UAT) Analysis on NeRV}
We provide more analysis and discussions of UAT analysis on the NeRV system. We define the problem that current NeRV systems are attempting to address and provide a comparison with existing video neural coding pipelines.

\subsection{Implicit Neural Video Coding Problem}\label{app_uat}
Following the pipeline of Implicit Neural Video Coding (INVC) presented in Sec.~\ref{invc_sec}
, we recall the proposed Implicit Neural Video Coding Problem (INVCP) as follows.
\begin{problem}(INVC Problem).\label{IVNCP}
The goal of Implicit Neural Video Coding is to find out the optimal design of the decoder $\mathcal{D}$ and encoder $\mathcal{E}$ in pursuit of minimal parameter quantity $\mathsf{Param}(\mathcal{D})$ and embeddings $ \{ e_{t}=\mathcal{E} \left( t \right) \in \mathbb{R}^{d_{in}^t} \}^{T}_{t=1}$ (where $d = d^{t}_{in}$ is often the same for all $t$ in existing NeRV systems) under a certain approximation error $\epsilon$ between the reconstruction $\tilde{V}$ and a given video sequence $V$,
\begin{gather}\label{IVNC}
\mathop{\arg\min}_{\mathcal{D}, \mathcal{E} }~ \mathsf{Param} \left( \mathcal{D} \right) + \sum_{t=1}^{T} d_{in}^t , \nonumber  \\
\mathrm{ s.t. } ~ L_{\mathcal{D}}, w_{\mathcal{D}} \in \left[1, \infty \right),~\mathop{\sup} \mathop{\sum} \Vert \tilde{V_t} - V_t\Vert \leq \epsilon,~t \in [1, T]. \notag
\end{gather}
\end{problem}
In the practice of INVC research, we usually use the dual problem of~\ref{IVNCP} to determine the optimal architecture of a model to achieve a certain level of accuracy for fitting the video. We name it the Dual Implicit Neural Video Coding Problem (DINVCP).
\begin{problem}(Dual INVC Problem).\label{DIVNCP}
Given a certain parameter quantity $\mu$, the Dual INVC problem aims to determine the optimal design of decoder $\mathcal{D}$ and encoder $\mathcal{E}$ to minimize the minimal approximation error between the reconstruction $\tilde{V}$ and the given video sequence $V$,
\begin{gather} \label{DIVNC}
\mathop{\arg\min}_{\mathcal{D}, \mathcal{E} }~\mathop{\sup} \mathop{\sum} \Vert \tilde{V_t} - V_t\Vert, \nonumber  \\
\mathrm{ s.t. } ~ L_{\mathcal{D}}, w_{\mathcal{D}} \in \left[1, \infty \right), \mathsf{Param} \left( \mathcal{D} \right) + \sum_{t=1}^{T} d_{in}^t \leq \mu ,~t \in [1, T].\notag
\end{gather}
\end{problem}

In practice, when using a NeRV model to represent a given video within a certain model size limit $\mu$ through end-to-end training, it is trying to solve the DINVCP.

\subsection{Comparison between DINVCP and Previous Neural Coding Pipelines}
Distribution-Preserving Lossy Compression (DPLC) is proposed by~\cite{Tschannen2018DeepGM} motivated by GAN-based image compression~\cite{Agustsson2018GenerativeAN}. It is defined as follows:
\[
\min_{E,D}~~ \mathbb{E}_{X,D} [d(X,D(E(X)))] + \lambda d_f(p_X,p_{\tilde{X}}),
\]
where $E, D, X, \tilde{X}$ are encoder, decoder, given input and reconstruction, $d_f$ is a divergence which can be estimated from samples. DPLC emphasizes the importance of maintaining distribution consistency for effective compression and reconstruction.

\cite{rdo} proposes Rate-Distortion Optimization (RDO). Later, \cite{Blau2017ThePT} reveals the importance of perceptual quality and proposes the Perception-Distortion Optimization (PDO) as
\[
\min_{p_{\tilde{X}|Y}}~ d(p_X, p_{\tilde{X}})~~s.t.~~ \mathbb{E}[\Delta(X, \tilde{X})] \leq D,
\]
where $\Delta$ is the distortion measure and $d$ is the divergence between distributions. Furthermore, \cite{Blau2019RethinkingLC} defines the Rate-Distortion-Perception Optimization (RDPO) as
\[
\min_{p_{\tilde{X}|X}} I(X, \tilde{X})~~s.t.~~ \mathbb{E}[\Delta(X, \tilde{X})] \leq D,~d(p_X, p_{\tilde{X}}) \leq P,
\]
where $I$ denotes mutual information. 


The primary objective, which also serves as the main obstacle in the aforementioned pipelines, is that density estimation is not only costly but also challenging to estimate accurately.
Different from DPLC, PDO, or RDPO, \textit{DINVCP does not need to model the distribution of the given signal explicitly.} In fact, the distribution of input images or videos is difficult to approximate. Whether it is approached by minimizing ELBO or through adversarial training~\cite{vae, gan}, there is always a certain gap or mismatch. Besides, other density estimation methods, such as flow-based or diffusion models, suffer from huge computational costs~\cite{ddpm, ldm}. In contrast, \textit{NeRV system implicitly models the unknown distribution of a given signal via specific decoding computation process under certain model parameter quantity constraints. The calculation process per se is regarded as the side information}~\cite{Wyner1976TheRF, Jonschkowski2015PatternsFL}. 

This approach of implicitly modeling distributions through computational processes under parameter quantity constraints aligns with some current perspectives that suggest the intelligence of Large Language Models (LLM) emerges from data compression~\cite{cagi, Deletang2023LanguageMI}. LLMs such as GPT aim to transfer as much data as possible to models of the same size for learning (and continue to increase the model size after learning) to achieve information compression and efficient information coding. However, the NeRV system strives to compress the model size as much as possible for a given video, emerging with robust representations with generalized capability.

The improvement of PNeRV in terms of perceptual quality confirms this conjecture. By upgrading the model structure and training with only MSE loss, PNeRV emerges better perceptual performance without having to estimate the signal's unobtainable prior distribution.

\subsection{Proof of Theorem 1}\label{app_a}
Following the definitions given in Sec.~\ref{uat_sec}
, the width $w$ of $\mathcal{N}$ is named as $\mathop{\max} d_i, \{ d_i \in \mathbb{N} \}^L_{i=1}$. Once the minimal width $w^* = w_{min} \left( d_{in}, d_{out}\right)$ is estimated by $d_{in}, d_{out}$, such that, for any continuous function $f: [0,1]^{d_{in}} \to \mathbb{R}^{d_{out}}$ with $\epsilon \geq 0$, there exists a $\mathcal{N}$ with input dimension $d_{in}$, hidden layer widths at most $w^*$, and output dimension $d_{out}$ that $\epsilon-$approximates $f$:
\[
\mathop{\sup}_{x \in [0,1]^ {d_{in}}} \Vert f \left(x \right) - \mathcal{N} \left(x \right) \Vert \leq \epsilon.
\]
The goal of Theorem~\ref{param}
is to determine the minimum parameter demand when $\epsilon-$approximates the implicit $\mathcal{F}$ which represents the given video. We recall Theorem~\ref{param}
as Theorem~\ref{param_app} as follows for better illustration.

\begin{theorem}\label{param_app}
For a cascaded NeRV system to $\epsilon$-approximate a video $V$ which is implicitly characterized by a certain unknown L-Lipschitz continuous function $\mathcal{F}: K \to \mathbb{R}^{d_{out}}$ where $K \subseteq \mathbb{R}^{d_{in}}$ is a compact set, then the upper bound of the minimal parameter quantity $\mathsf{Param}(\mathcal{D})$ is given as\vspace{-0.02in}
\[
 \mathsf{Param}_{\min}(\mathcal{D})\leq d_{out}^2 \left( \frac{ \mathop{O}
    \left(diam  \left(K \right)
    \right) }{\omega_\mathcal{F}^{-1} \left( \epsilon \right) }
\right)^{d_{in}+1}.
\]
\end{theorem}

Before we start, we will recall the setup and demonstrate some mathematic concepts and lemmas.
\begin{definition}\label{string}
A function $g: \mathbb{R}^{d_{in}} \to \mathbb{R}^{d_{out}}$ is a max-min string of length $L \geq 1$ on $d_{in}$ input variables and $d_{out}$ output variables if there exist affine functions ${\ell}_1, \ldots ,{\ell}_L: \mathbb{R}^{d_{in}} \to \mathbb{R}^{d_{out}}$ such that
\[
g = \sigma_{L-1} ( \ell_L, \sigma_{L-2} \left(  \ell_{L-1},\cdots , \sigma_2 \left( \ell_3, \sigma_1\left( \ell_1, \ell_2\right)\right) \cdots \right).
\]
\end{definition}

The definition of max-min string and DMoC (Def.~\ref{moc}
) are first introduced in~\cite{UniversalFA} and~\cite{ApproximatingCF}. We introduce two lemmas, which were presented as Propositions {\color{red}2} and {\color{red}3} in~\cite{ApproximatingCF}. 

\begin{lemma}\label{L2}~\cite{ApproximatingCF}
For every compact $K \subseteq \mathbb{R}^{d_{in}}$, any continuous $f : K \to \mathbb{R}^{d_{out}}$ and each $\epsilon \geq 0$, there exists a max-min string $g$ on $d_{in}$ input variables and $d_{out}$ output variables with length
\[
\left( \frac{ \mathop{O}
    \left(diam  \left(K \right)
    \right) }{\omega_f^{-1} \left( \epsilon \right) }
\right)^{d_{in}+1},
\]
for which
\[
\Vert f-g \Vert_{C^0(K)} \leq \epsilon.
\]
\end{lemma}

\begin{lemma}\label{L1}~\cite{ApproximatingCF} 
For every max-min string $g$ on $d_{in}$ input variables and $d_{out}$ output variables with length $L$ and every compact $K \subseteq \mathbb{R}^{d_{in}}$, there exists a \textsc{ReLU} net $\mathcal{N}$ with input dimension $d_{in}$, hidden layer width $d_{in}+d_{out}$, and depth $L$ that computes $x \mapsto g(x)$ for every $x \in K$.
\end{lemma}

\begin{lemma}\label{L3}
~\cite{MinimumWF}
For any $p \in [1, \infty )$, \textsc{ReLU} nets of width $w$ are dense in $L^P (\mathbb{R}^{d_{in}}, \mathbb{R}^{d_{out}})$ if and only if $w \geq \max\{d_{in} + 1, d_{out}\}$.
\end{lemma}\vspace{-0.05in}

The proofs of Lemma~\ref{L2} and \ref{L1} can be found in the Sec.~{\color{red}2.1} and Sec.~{\color{red}2.2} of \cite{ApproximatingCF}. Lemma~\ref{L3} is the Theorem 1 demonstrated in \cite{MinimumWF} with its proof. Now we provide the proof of Theorem~\ref{param_app} as follows.

\begin{proof}
From Lemma~\ref{L2}, the implicit function $\mathcal{F}_{\mathcal{V}}$ which represents the video $\mathcal{V}$ can be approximated by one max-min string $g$. It is worth mentioning that $\mathcal{F}_{\mathcal{V}}$ is supposed to be continuous because video can be considered as a slice of the real world. The length of this max-min string $g$ is given by Lemma~\ref{L2}. According to Lemma~\ref{L1}, there exists a \textsc{ReLU} net $\mathcal{N}_g$ with the same input and output dimensions that fit this max-min string. So, the minimal parameters of $\mathcal{N}_g$, also the sum of weights for each layer, is
\[
\mathbf{Param} =\sum_{l=1}^{L} w_l w_{l-1},
\]
where $w_l$ is width in each hidden layer and $L$ is given in Lemma~\ref{L2}. Noticed that the \textit{whole width $w$} of a model is the upper bound of all hidden layer widths ${ \{ w_l \} }^L_{l=0}$. $w_{\min}$ is the minimum estimate for this upper bound, $w_l \leq w_{\min} \leq w$. $w_{\min}$ is further contracted from $d_{in}+ d_{out}$ to $\max\{d_{in}+1, d_{out} \}$ by~\cite{MinimumWF} (Lemma~\ref{L3}).

Thus, the minimal parameters of $\mathcal{N}_g$ under a certain error is no longer than
\begin{equation}
\begin{split}
\mathbf{Param}_{\min} & \leq w_{\min}^2 \left( \frac{ \mathop{O}
    \left(diam  \left(K \right)
    \right) }{\omega_f^{-1} \left( \epsilon \right) }
\right)^{d_{in}+1} \nonumber \\
& = d_{out}^2 \left( \frac{ \mathop{O}
    \left(diam  \left(K \right)
    \right) }{\omega_f^{-1} \left( \epsilon \right) }
\right)^{d_{in}+1}, \nonumber
\end{split}
\end{equation}
where $w_{\min}= d_{out}$ for video $\mathcal{V}: \mathbb{N}\to \mathbb{R}^{d_{out}}$. Equality is reached when each layer width reaches the upper bound of minimal width, the worst case.
\end{proof}

Although the upper bound of $\mathsf{Param}(\mathcal{D})$ is fixed regardless of the detailed architecture, the actual performance of serial NeRV will be influenced by structure design, parameter initialization, activation functions, loss functions, and optimizer.

\section{More Related Works}\label{supp_rw}

\noindent\textbf{Comparison with Other Subpixel-based Upsampling Operators}. The NeRV system aims at reconstructing high-resolution videos through decoding low-dim embeddings. Therefore, proper upsampling operators are crucial for its performance. Existing subpixel-based upsampling operators are not efficient enough for the NeRV system. Deconv~\cite{deconv} pads the subpixels with zeros and passes them through a Conv layer, resulting in block artifacts~\cite{deconv_arti}. PixelShuffle~\cite{ps} first expands the feature map channels through a \textsc{Conv} and then rearranges them into the target subpixels. However, the desired subpixels of a given position are only related to the expanding channels of the same position, ignoring contextual information, as shown in Fig.~\ref{ps_kfc} of the main text. Additionally, PixelShuffle encounters an exponential explosion of required channels when the upsampling ratio is large.

\noindent\textbf{Comparison with INR on Images}.~\cite{siren} (SIREN) uses sine as a periodic activation function to model the high-frequency information of a given image~\cite{fourier} and performs a sinusoidal transformation before input~ \cite{Xu_2022_INSP} tries to directly modify an INR without explicit decoding. The main difference between these methods and ours is that we consider the input coordinate-pixel pairs to be \textit{dense} for the INR on image coding. In a natural image, the RGB value at a specific position is often closely related to its neighboring positions. However, for high-resolution videos, the gap between adjacent frames can be much larger, both in terms of pixels and semantic terms. This situation is akin to only observing partial pixels from a given image.

\noindent\textbf{Comparison with Self-attention Module}. Self-attention (SA) and Multi-head Self-attention (MSA) modules~\cite{NonlocalNN, AttentionIA, SwinTH, t2023is} compute the response at a position by attending to all positions, which is similar to \textsc{KFc}. The major defect of SA and MSA when adopted in NeRV is that the computational complexity and the space complexity are too high to efficiently compute the global correlations between arbitrary positions, especially the computational cost ($O(n^2 d)$) between queries and keys for high-resolution feature maps. \textsc{KFc} not only captures long-range dependencies but also achieves low-cost rescaling, both of which are significant for NeRV.

\section{Additional Results}
Unless otherwise specified, all models utilized in the additional results are trained on a 3M model for 300 epochs.
\subsection{Comparison of Generalization Ability by Video Interpolation Results}\label{app_int}
Indeed, the concepts of approximation and generalization are distinct topics within the field of deep learning theory~\cite{Nakkiran2019DeepDD, Advani2017HighdimensionalDO}. Understanding the causal relationship between overfitting and the generalization capacity of NeRV necessitates further investigation. Existing NeRV models always focus on the models' approximation capabilities through overfitting training. 

Nonetheless, we also evaluate the generalization performance of our proposed PNeRV through a video interpolation experiment. Adhering to the experimental methodology employed in \cite{HNeRV} and \cite{DNeRV}, the model is trained using odd-numbered frames and then tested with unseen even-numbered frames. The results, presented in Table~\ref{int}, indicate that PNeRV surpasses most baseline methods. Future research will focus on the theoretical analysis and enhancement of PNeRV's generalization abilities.

\begin{table*}[h]
    \setlength{\tabcolsep}{5pt}
  \centering\small
      \setlength{\abovecaptionskip}{0.05in}
      \setlength{\belowcaptionskip}{-0.05in}
  \begin{tabular}{l||lllllll|c}
    & Beauty & Bospho & Honey & Jockey & Ready & Shake &Yacht &Avg.\\
    \midrule[1.5pt]
    NeRV~\cite{nerv}     &28.05  & 30.04 & 36.99  &20.00 &17.02  & 29.15  &24.50 & 26.54 \\
    E-NeRV~\cite{enerv}  &27.35  & 28.95 & 38.24  &19.39 &16.74  & 30.23  &22.45 & 26.19 \\
    H-NeRV~\cite{HNeRV}  &31.10  & 34.38 & 38.83  &23.82 &20.99  & 32.61  &27.24 & 29.85 \\
    DiffNeRV~\cite{DNeRV}&35.99  & 35.10 & 37.43  &30.61 &24.05  & 35.34  &28.70 & \textbf{32.47} \\
 \midrule
    PNeRV                &33.64  & 34.09 & 39.85  &28.74 &23.12  & 31.49  &27.35 & \underline{31.18} \\
  \end{tabular}
    \caption{Video interpolation results on 960 $\times$ 1920 UVG in PSNR.}
  \label{int}
\end{table*}

\begin{table*}[!t]
    \setlength{\tabcolsep}{3pt}
  \centering
  \begin{tabular}{l||cccccccc|c}
     & Bmx-B  & Camel & Dance-J & Drift-C & Elephant & Parkour &Scoo-G &Scoo-B &Avg.\\
    \midrule[1.5pt]
    HNeRV        &20.39  &21.85   &21.73  &28.81  &17.35  &19.97  &\textbf{24.49}  &19.76  &21.79    \\
    DiffNeRV     &\textbf{22.95}  &23.72   &21.78  &\textbf{30.37}  &26.02  &21.55  &22.78  &21.00  &23.77\\
\midrule
\textbf{PNeRV}   &21.69  &\textbf{24.28}  &\textbf{25.21}  &30.01  &\textbf{27.32}  &\textbf{22.61}  &22.84  &\textbf{22.61}  &\textbf{24.57}\\
  \end{tabular}
  \vspace{-0.1in}
  \caption{Video inpainting results using center mask on 960 $\times$ 1920 DAVIS in PSNR.}
  \label{inp_c_p}
\end{table*}

\begin{table*}[!t]
    \setlength{\tabcolsep}{3pt}
  \centering
  \begin{tabular}{l||cccccccc|c}
          & Bmx-B  & Camel & Dance-J & Drift-C & Elephant & Parkour &Scoo-G &Scoo-B &Avg.\\
    \midrule[1.5pt]
    HNeRV        &0.665  &0.733   &0.677  &0.650  &0.489  &0.650  &0.859  &0.789  & 0.725    \\
    DiffNeRV     &0.767  &0.815   &0.667  &\textbf{0.949}  &0.817  &0.754  &0.852  &\textbf{0.844} &0.808\\
\midrule
\textbf{PNeRV}   &\textbf{0.802} &\textbf{0.844}  &\textbf{0.792}  & 0.947  &\textbf{0.862}  &\textbf{0.801}  &\textbf{0.874}  &0.812  &\textbf{0.842}\\
  \end{tabular}
  \vspace{-0.1in}
  \caption{Video inpainting results using center mask on 960 $\times$ 1920 DAVIS in SSIM.}
  \label{inp_c_s}
\end{table*}

\begin{table*}[!t]
    \setlength{\tabcolsep}{3pt}
  \centering
  \begin{tabular}{l||cccccccc|c}
       & Bmx-B  & Camel & Dance-J & Drift-C & Elephant & Parkour &Scoo-G &Scoo-B &Avg.\\
    \midrule[1.5pt]
    HNeRV        &23.16  &20.94   &26.54  &31.70  &17.36   &21.32  &\textbf{26.89}  &21.05  &23.62     \\
    DiffNeRV     &\textbf{25.70}  &\textbf{24.71}   &26.59  &34.74  &25.93 &24.51  &26.61  &\textbf{24.27}  &\textbf{26.63}\\
\midrule
\textbf{PNeRV}   &24.96   &24.18  &\textbf{26.62}  &\textbf{34.84}  &\textbf{27.50}  &\textbf{24.98}  &26.85  &22.13  &26.51\\
  \end{tabular}
  \vspace{-0.1in}
  \caption{Video inpainting results using disperse mask on 960 $\times$ 1920 DAVIS in PSNR.}
  \label{inp_d_p}
\end{table*}

\begin{table*}[!t]
    \setlength{\tabcolsep}{3pt}
  \centering
  \begin{tabular}{l||cccccccc|c}
      & Bmx-B  & Camel & Dance-J & Drift-C & Elephant & Parkour &Scoo-G &Scoo-B &Avg.\\
    \midrule[1.5pt]
    HNeRV        &0.728  &0.661  &0.779 &0.957  &0.490   &0.685  &0.889  &0.794  &0.748     \\
    DiffNeRV     &0.819  &0.832   &0.795  &0.972  &0.827 &0.799  &0.892  &\textbf{0.897}  &0.854\\
\midrule
\textbf{PNeRV}   &\textbf{0.843}   &\textbf{0.854} &\textbf{0.806}  &\textbf{0.975}  &\textbf{0.877}  &\textbf{0.836}  &\textbf{0.910} &0.866  &\textbf{0.871}\\
  \end{tabular}
  \vspace{-0.1in}
  \caption{Video inpainting results using disperse mask on 960 $\times$ 1920 DAVIS in SSIM.}
  \label{inp_d_s}
\end{table*}

\begin{figure}[t]
  \centering
  	\begin{minipage}{1\linewidth}
     \setlength{\belowcaptionskip}{-0.05in}
		\centering
  \includegraphics[width=0.8\linewidth]{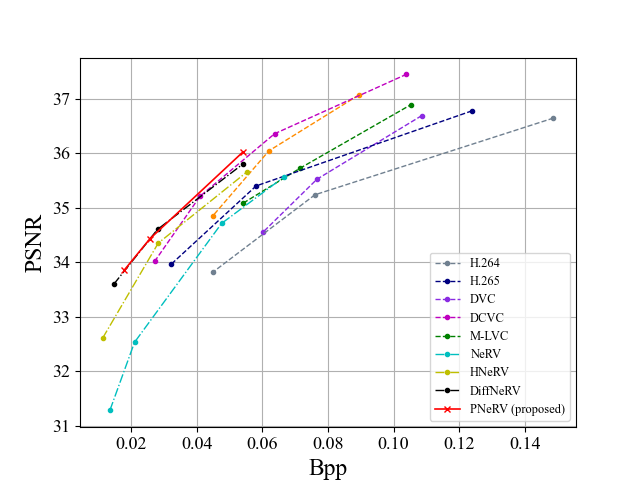}
  \caption{PSNR of video compression on UVG.}
  \label{comp_p}
   \end{minipage}
   \hfill
  \begin{minipage}{1\linewidth}
  \centering
  \includegraphics[width=0.8\linewidth]{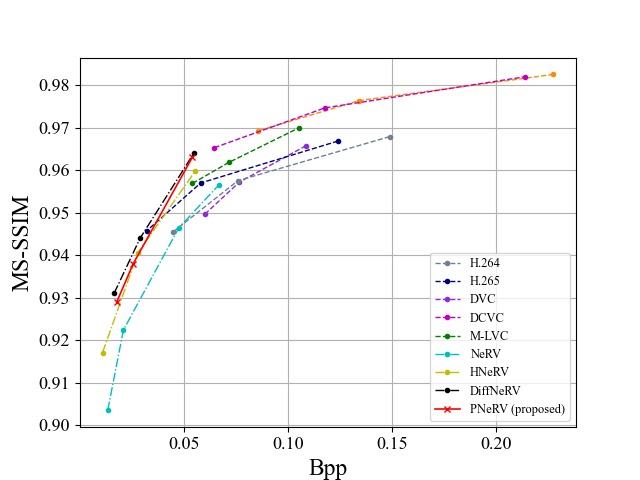}
  \caption{SSIM of video compression on UVG.}
  \label{comp_s}
   \end{minipage}
\end{figure}

\begin{figure}[h]
    \centering
         \setlength{\belowcaptionskip}{-0.05in}
  \includegraphics[width=0.8\linewidth]{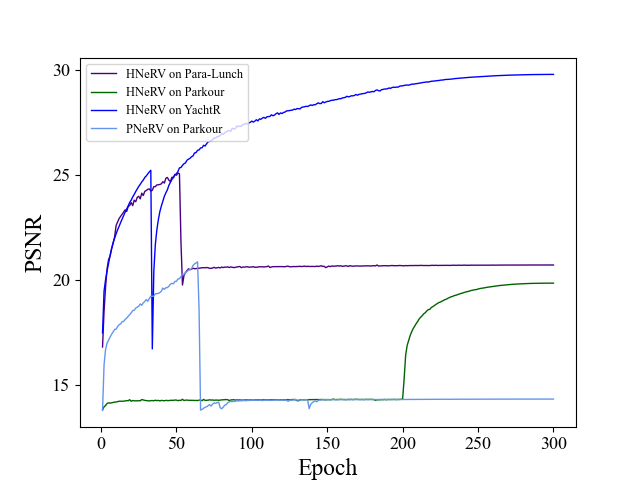}
\vspace{-0.05in}\caption{Example of training difficulty of different NeRV methods in 3M size.}\label{train}
\end{figure}

\subsection{Comparison of Video Compression and Discussion of Training Difficulties}\label{app_comp}
The video compression comparison of PNeRV with other NeRV models in terms of PSNR and MS-SSIM is shown in Fig.~\ref{comp_p} and Fig.~\ref{comp_s}. Following the same settings utilized in~\cite{HNeRV, DNeRV}, we evaluate the video compression comparison with 8-bit quantization for both embeddings and the model without model pruning. 

PNeRV has demonstrated remarkable performance, notably outperforming conventional encoding pipelines like H264~\cite{h264} and H265~\cite{h265}, and possesses substantial advantages over several traditional neural video coding models~\cite{dvc, MLVC, dcvc}, particularly at low bit rates. Compared to INR-based methods, PNeRV has also achieved competitive results and outperforms other NeRV methods~\cite{nerv, HNeRV, DNeRV} in terms of PSNR. 

For detailed experimental settings, PNeRV adjusts the size of the decoder and the dimensions of the input diff embedding to validate the encoding performance of the proposed method across various bit rates. At low bit rates, the encoding performance of the model may experience some degradation. We believe this is due to the diversity and complexity of the modules required by PNeRV. Maintaining a certain amount of parameters (such as the number of channels in convolutional layers) is crucial for preserving performance. This ensures that the model has sufficient capacity to handle the challenges posed by low-bit rate encoding.

It is worth noting that all implicit models encounter significant training challenges when dealing with large parameters, such as those exceeding 5M. As a result, these models often converge to local minima, leading to trivial outputs. This issue poses a significant obstacle to the compression performance of all NeRV methods, particularly when the Bpp value increases. Some examples of training failure are shown in Fig.~\ref{train}, where models are 3M under the same conditions.

\subsection{Comparison of Robustness by Video Inpainting Results}\label{app_inp}
We evaluate the robustness of different methods using video inpainting tasks following the same setting as in \cite{HNeRV} and \cite{DNeRV}, which use a center mask and disperse mask. The center mask uses a rectangular area that occupies one-fourth of the width and height of the original frame, positioned at its center. The disperse mask comprises five square areas, each measuring $100 \times 100$ pixels, positioned in the four corners and the center of the frame. The pixel value of areas in the masks is reset to 0. The trained models in video regression tasks will be directly utilized for inpainting without any fine-tuning. Models take the masked frames as input and try to predict the original ones.

The results using the center mask are provided in Tab.~\ref{inp_c_p} and Tab.~\ref{inp_c_s}. The dispersed ones are in Tab.~\ref{inp_d_p} and Tab.~\ref{inp_d_s}. PNeRV acquires competitive results with both the center mask and the disperse mask, indicating robust modeling capability.

\subsection{More Visualization Examples for Perceptual Quality}\label{app_qual}
We show some more examples of qualitative comparisons between different models.

\begin{figure*}[t]
  \centering
  \includegraphics[width=0.95\linewidth]{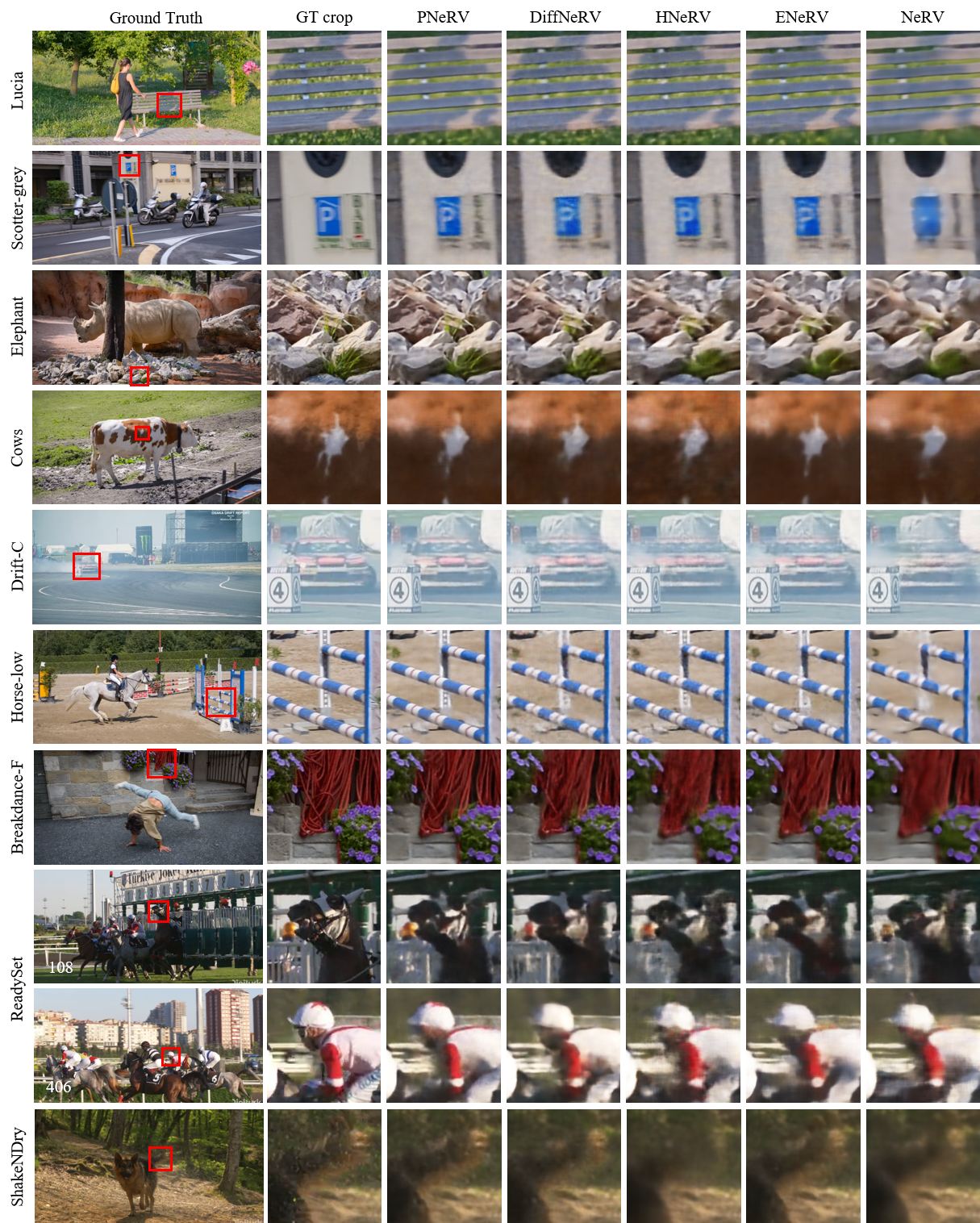}\\
  \vspace{-0.05in}\caption{Visual comparison examples on various videos.}
  \label{add_davis}
\end{figure*}

Shown in Fig.~\ref{add_davis}, the results of PNeRV are smoother and less noisy. For instance, in ``Lucia'' and ``Horse-low'', PNeRV pays more attention to the geometric pattern of the main objects and ignores those high-frequency details of the background scene. Other baseline methods cannot reconstruct frames at such a semantic level. Due to the lack of high-level information guidance and a global receptive field, baseline methods are hard to reasonably allocate model weights to more important objects, e.g., red waterpipe in ``Breakdance-flare'' and patterns in ``Cows''.

\begin{figure*}[t]
  \centering
  \includegraphics[width=0.95\linewidth]{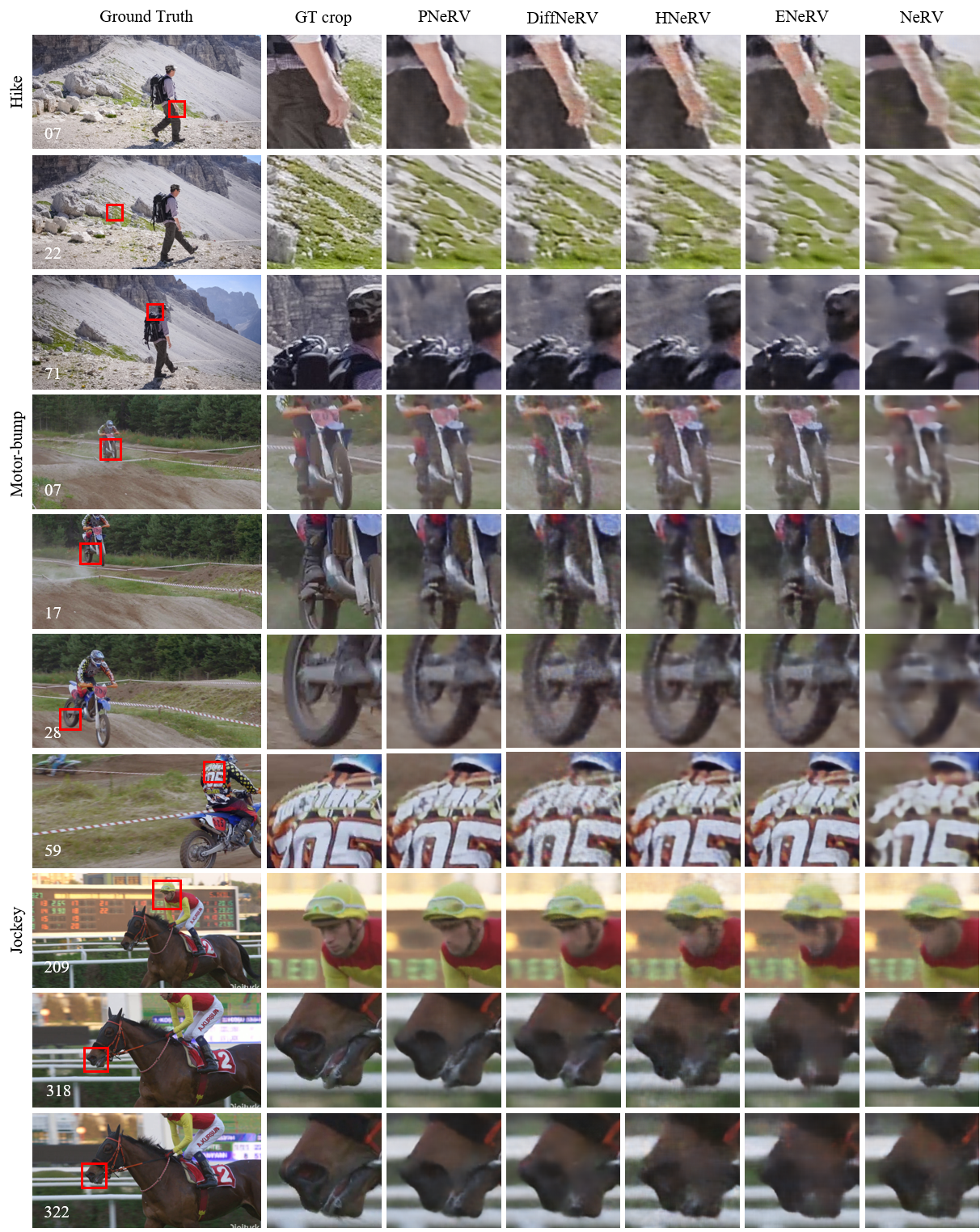}\\
  \vspace{-0.05in}\caption{Visual comparison examples on the same video by same models. Corresponding time stamps are shown in the bottom left.}
  \label{same_davis}
\end{figure*}

Shown in Fig.~\ref{same_davis}, the comparison at different timestamps of the same video indicates some specific common issues of different models.
Overlapping and noisy patterns have occurred in the results of DiffNeRV~\cite{DNeRV} and HNeRV~\cite{HNeRV}, such as the grass and hands in ``Hike''. ENeRV~\cite{enerv} and NeRV~\cite{nerv} often result in color deviation and blurring, e.g., backpack in ``Hike'' and motor in ``motor-bump''. PNeRV achieves a balance between preserving details and maintaining semantic consistency. Compared to DiffNeRV, which also uses the difference between frames as input, the latter's reconstruction of details is unbiased. However, human attention to visual elements under different semantics should be different. Improving the reconstruction results through high-level information is one of PNeRV's pursuits.

\subsection{Discussion on the Failure Cases}
As shown in Table~\ref{davis}
, PNeRV fails in the ``Dog'' which is blurred and mixed with jitter and deformation. Also, the ``Soapbox'' video, which comprises two clips from entirely different scenes connected by a few frames where the camera rotates through a large angle, poses a challenge. So far, PNeRV has not been able to handle severe temporal inconsistency effectively.

\subsection{Video Examples}\label{v_examples}
We provide some video examples from DAVIS as follows. From the video comparison, it can be seen that the reconstructions of NeRV have lost spatial details, and it is difficult for DNeRV to reconstruct videos containing pervasive scattered high-frequency details. Whether there is large motion or high-frequency details in the given videos, PNeRV is more robust in modeling the spatial consistency, leading to better perceptual quality in reconstructions. The links to the examples are presented as follow.\\
Dance-jump: \url{https://drive.google.com/file/d/18JZq1BCkBJWCkZs-71OB7wI6j_Vma0vP/view?usp=drive_link}\\
Elephant: \url{https://drive.google.com/file/d/1rnPEsEtfA5UADU6BnwEDOPRG9hO9uPuM/view?usp=drive_link}\\
Kite-surf: \url{https://drive.google.com/file/d/1DDGw1zc2iJWcJHdBS4DOnfUQVf2H04Bs/view?usp=drive_link}\\
Parkour: \url{https://drive.google.com/file/d/1jWbJuoc-GCz2N_dXAJSER0PSy7ThrMr-/view?usp=drive_link}\\
Scooter-grey: \url{https://drive.google.com/file/d/1vs22Ru-AwAQuG710qbF72lwdHS1ABy83/view?usp=drive_link}

\begin{table*}[!t]\footnotesize
      \centering
      \setlength{\abovecaptionskip}{0.05in}
      \setlength{\belowcaptionskip}{-0.05in}
       \tabcolsep=0.09cm
  \centering
  \begin{tabular}{l||lllllll|c|c}
     Models                   &Bmx-B & Camel & Dance-J & Dog & Drift-C & Parkour &Soapbox	 &Avg. &A.P.G \\
    \midrule[1.2pt]
    NeRV~\cite{nerv}  &29.42/0.864  &24.81/0.781  &27.33/0.794  &28.17/0.795 &36.12/0.969 &25.15/0.794 &27.68/0.848 &28.38/0.835 &-\\
    E-NeRV~\cite{enerv} &28.90/0.851  & 25.85/0.844 & 29.52/0.855 &30.40/0.882 &39.26/0.983  &25.31/0.845 &28.98/0.867 &29.75/0.875 &-\\
    HNeRV~\cite{HNeRV}  &29.98/0.872 & 25.94/0.851 &29.60/0.850 &30.96/0.898 &39.27/0.985  &26.56/0.851  &29.81/0.881 & 30.30/0.874 &-\\
    DiffNeRV~\cite{DNeRV} &30.58/0.890  & 27.38/0.887 &29.09/0.837 &\textbf{31.32/0.905} &40.21/0.987 & 25.75/0.827 &\textbf{31.47/0.912} &30.84/0.892 &-\\
 \midrule[1.2pt]
    \textit{Ablation Study} \\
 \midrule[1.2pt]
    Bilinear + Concat &24.85/0.783 &24.49/0.793 &28.32/0.806 &26.19/0.723 &31.92/0.943 &25.09/0.793 &29.23/0.872 &27.16/0.816 &\textcolor{red}{-4.07}\\
    Bilinear + GRU &29.86/0.874 &25.00/0.811 &29.16/0.830 &27.11/0.753 &32.09/0.945 &26.43/0.845 &29.10/0.874 &28.39/0.847 &\textcolor{red}{-2.84}\\
   Bilinear + LSTM &26.22/0.792 &26.87/0.871 &27.85/0.788 &26.71/0.741 &33.65/0.946 &25.82/0.820 &29.42/0.881 &28.07/0.834 &\textcolor{red}{-3.16}\\
    Bilinear + \textbf{BSM} &29.97/0.877 &27.35/0.881 &29.49/0.838 &27.14/0.756 &34.34/0.968 &26.15/0.835 &29.14/0.876 &29.08/0.862  &\textcolor{red}{-2.15}\\
 \midrule[0.1pt]
    DeConv + Concat &28.06/0.840 &24.07/0.774 &27.86/0.792 &25.16/0.693 &34.97/0.961 &22.13/0.683 &29.33/0.877 &27.37/0.803  &\textcolor{red}{-3.86}\\
    DeConv + GRU &27.52/0.827 &28.16/0.900 &29.09/0.825 &25.76/0.706 &37.91/0.980 &25.09/0.793 &29.54/0.882 &29.00/0.845  &\textcolor{red}{-2.23}  \\
   DeConv + LSTM &30.15/0.882 &26.49/0.859 &28.30/0.805 &25.94/0.712 &34.91/0.956 &26.35/0.842 &30.26/0.895 &28.91/0.850 &\textcolor{red}{-2.32}\\
   DeConv + \textbf{BSM} &31.56/0.906 &27.18/0.878 &29.77/0.847 &30.09/0.868 &36.03/0.971 &26.09/0.831 &29.00/0.872 &29.96/0.881 &\textcolor{red}{-1.27}\\
  \midrule[0.1pt]
  \textbf{KFc} + Concat &27.51/0.826 &25.02/0.816 &29.02/0.831 &28.80/0.831 &36.82/0.974 &25.12/0.796 &28.53/0.864 &28.68/0.848 &\textcolor{red}{-2.55}\\
  \textbf{KFc} + GRU &\textbf{31.69/0.910} &25.88/0.848 &28.32/0.805 &28.47/0.813 &33.25/0.942 &26.68/0.853 &30.89/0.903 &29.31/0.868 &\textcolor{red}{-1.92}\\
 \textbf{KFc} + LSTM &29.16/0.862 &27.24/0.878 &28.90/0.825 &29.28/0.842 &32.73/0.935 &26.62/0.839 &29.35/0.879 &29.04/0.866 &\textcolor{red}{-2.19}\\ 
 \textbf{KFc} + \textbf{BSM} (\textbf{PNeRV})   &31.05/0.896  & \textbf{27.89/0.892} & \textbf{30.45/0.873} &31.08/0.898 &\textbf{40.23/0.987}  &\textbf{27.08/0.867} &30.85/0.902 & \textbf{31.22/0.902} & \textcolor{blue}{+0}\\
  \end{tabular} 
   \caption{Ablation results on DAVIS subset in PSNR and MS-SSIM, where Avg. is the average PSNR and A.P.G is the average PSNR gap. Every result is reported by corresponding model trained in 300 epoch and 3M size.
 }  \label{davis_abla_detail}
\end{table*}

\begin{table*}[!t]
	\noindent
     \setlength{\abovecaptionskip}{0.05in}
     \setlength{\belowcaptionskip}{-0.05in}
	\begin{minipage}{0.3\textwidth}
		\centering
  \resizebox{!}{0.7cm}{
		\begin{tabular}{c||ccc}
                & 40$\times$80  & 20$\times$40 & 10$\times$20 \\
        \midrule[1.2pt]
        PSNR    &\textbf{31.94} & 31.33 &30.50\\
        SSIM &\textbf{0.960} & 0.954 &0.947\\
        \end{tabular}
        }
		\caption{Embedding size in PNeRV-L.}
		\label{abla_1}
         \end{minipage}
	\begin{minipage}{0.3\textwidth}
		\centering
	  \resizebox{!}{0.7cm}{
		\begin{tabular}{c||ccc}
         & 1$\times$1 & 3$\times$3  & 5$\times$5  \\
        \midrule[1.2pt]
        PSNR    &31.92 & \textbf{31.94}& 31.90\\
        SSIM &0.960 &0.960 & \textbf{0.961} \\
         \end{tabular}
         }
       	\caption{Kernel size in BSM.}
		\label{abla_2}
	\end{minipage}
 	\begin{minipage}{0.3\textwidth}
		\centering
  \resizebox{!}{0.7cm}{
 \begin{tabular}{c||ccc|c}
               & ReLU & Leaky & GeLU & w/o BN\\
        \midrule[1.2pt]
        PSNR &    31.80 & 31.86 & \textbf{31.94} &31.53\\
        SSIM &0.959& \textbf{0.961} &0.960 & 0.959\\
         \end{tabular}
         }
		\caption{Activation and BN in \textsc{KFc}.}
		\label{abla_3}
         \end{minipage}
\end{table*}

\section{Additional Ablation Studies}

\subsection{Ablation Results of Model Structure Details}\label{app_abla}
We ablate the structure details of PNeRV in 3M on ``Rollerblade'' in $480 \times 960$ from DAVIS, given in Tab.~\ref{abla_1}, Tab.~\ref{abla_2} and Tab.~\ref{abla_3}.
The alternation of kernel size or activation has little influence. Encoding more information into embeddings will help the decoder reconstruct better and also increase the overall size.

\subsection{Ablation Results of Proposed Modules on DAVIS}\label{app_dav}
To verify the contribution of different modules in PNeRV, we conduct ablation studies on (1) upscaling operators and (2) gated memory mechanisms. We compare KFc with two upscaling layers, Deconv and Bilinear, where ``Deconv'' is implemented by ``nn.ConvTranspose2d'' from PyTorch, and ``Bilinear'' is the combination of bilinear upsampling and Conv2D. KFc achieves better performance due to the global receptive field regardless of what fusion module it is combined with. 

Also, to illustrate the importance of adaptive feature fusion and improvement of BSM, we compare BSM with Concat, GRU and LSTM, where ``Concat'' means directly concatenating two features from different domains together. The ablation results suggest that the adaptive fusion of features from different domains significantly improves performance, and BSM outperforms other memory cells due to the disentangled feature learning. The last row is the final PNeRV and the last column shows PSNR gaps when changing modules in PNeRV.

\subsection{Visualization of Feature Maps}\label{app_fea}
\begin{figure*}[t]
  \centering
  \includegraphics[width=1\linewidth]{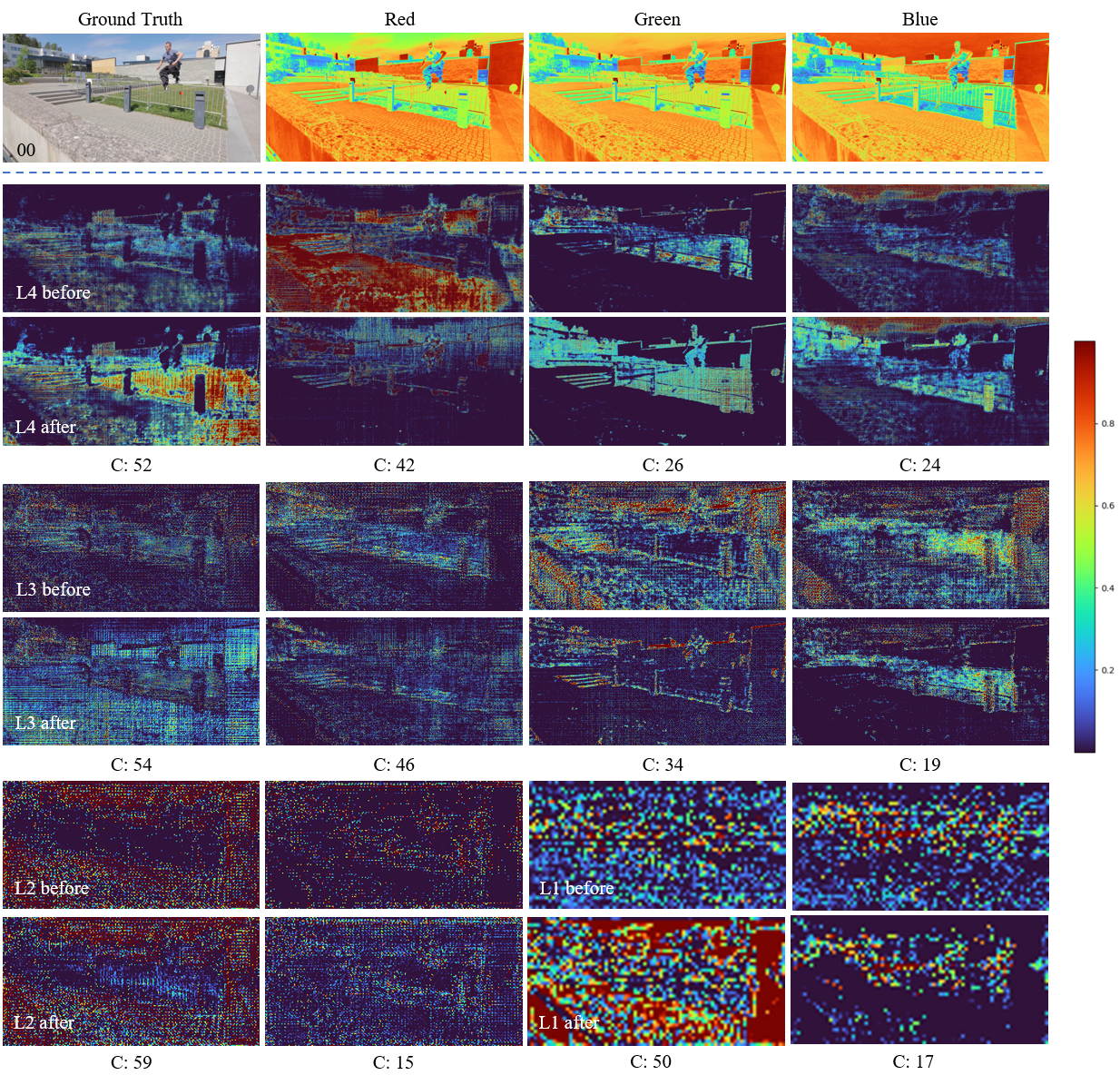}\\
  \vspace{-0.05in}\caption{Visualization examples of feature maps in different layers.``C'' refers to the channel number and ``L'' is the layer number.}
  \label{bsm_1}
\end{figure*}

\begin{figure*}[t]
  \centering
  \includegraphics[width=1\linewidth]{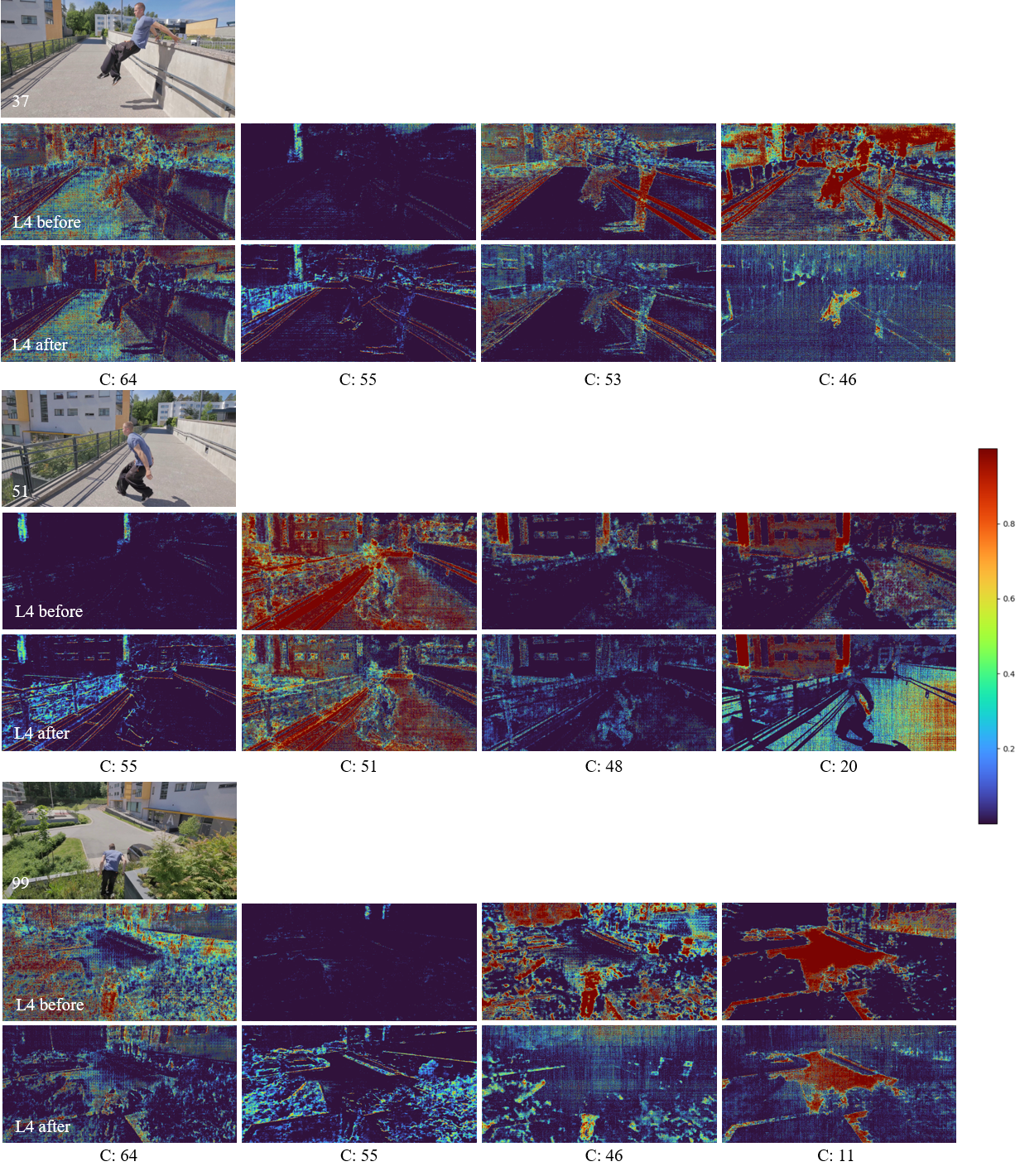}\\
  \vspace{-0.05in}\caption{Visualization examples of feature maps for different frames.``C'' refers to the channel number and ``L'' is the layer number.}
  \label{bsm_2}
\end{figure*}

To verify the effectiveness of hierarchical information merging via \textsc{KFc} and BSM, we visualize some feature maps in PNeRV-L which was pretrained on ``Parkour'' as examples. Those feature maps shown in Fig.~\ref{bsm_1} are from different channels and layers using the same frame as input. Those in Fig.~\ref{bsm_2} are all from the $4$-th layer but using different frames as input. The feature maps from $4$-th layer are in $480 \times 960$, and the original frames are in $960 \times 1920$. For each lower layer, the height and width are halved compared to the upper layer. ``Before'' and ``After'' refer to the feature maps before and after passing through BSM or after.

Fig.~\ref{bsm_1} illustrates how the coarse features are refined by BSM. Different channels respond to distinct spatial patterns of video frames, including factors like color, geometric structure, texture, brightness, motion, and so on. Before being processed by the BSM, the vanilla features are semantically mixed and entangled. However, the BSM is able to decouple these features and distinguish their specific effects, resulting in more refined and distinct outputs.

Additionally, for imperfect feature maps, BSM can add details or balance the focus of the reconstruction across various areas in the frames. These phenomena are commonly observed in the $4$-th layer, which is responsible for preparing for fine-grained reconstruction, as demonstrated in Figure.~\ref{bsm_2}. This shows the effectiveness of BSM in enhancing the quality of feature maps and improving the overall reconstruction.

\end{document}


\appendix
\section*{Appendix}
The appendix includes a discussion on related work, additional results, additional ablation studies, proof of Theorem~\ref{param} and pseudocode. The references number may be different from the main text.
\section{Discussion on related work}

\textbf{Compare to previous neural coding theories}. Distribution-preserving lossy compression (DPLC) is proposed by~\cite{Tschannen2018DeepGM} motivated by GAN-based image compression~\cite{Agustsson2018GenerativeAN}, and is given as
\[
\min_{E,D}~~ \mathbb{E}_{X,D} [d(X,D(E(X)))] + \lambda d_f(P_{\tilde{X}},P_X),
\]
where $E, D, X, \tilde{X}$ are encoder, decoder, given input and reconstruction, $d_f$ is a divergence which can be estimated from samples. DPLC believes that it is important to keep distribution consistent for compression and reconstruction. \cite{Blau2017ThePT} reveals the importance of perceptual quality and propose the perception-distortion optimization as
\[
\min_{p_{\tilde{X}|Y}}~ d(p_X, p_{\tilde{X}})~~s.t.~~ \mathbb{E}[\Delta(X, \tilde{X})] \leq D,
\]
where $\Delta$ is distortion measure and $d$ is divergence between distributions. Further, \cite{Blau2019RethinkingLC} defines the rate-distortion-perception optimization as
\[
\min_{p_{\tilde{X}|X}} I(X, \tilde{X})~~s.t.~~ \mathbb{E}[\Delta(X, \tilde{X})] \leq D,~d(p_X, p_{\tilde{X}}) \leq P,
\]
where $I$ denotes mutual information. Recall that Dual INVC (DINVC) problem is
\begin{gather} \label{DIVNC}
\mathop{\arg\min}_{\mathcal{D}, \mathcal{H} }~\mathop{\sup} \Vert \mathcal{G}(t) - \mathcal{V}(t) \Vert, \nonumber  \\
\mathrm{ s.t. } ~ L_{\mathcal{D}}, w_{\mathcal{D}} \in \left[1, \infty \right), \mathsf{Param} \left( \mathcal{D} \right) + \sum_{t=1}^{T} d_{in}^t \leq p ,~t \in [1, T],
\end{gather}
where $\mathcal{V}$ is given video, $\mathcal{G}=\mathcal{D} \circ \mathcal{H}$ is NeRV system consisting of encoder $\mathcal{H}$ and decoder $\mathcal{D}$ which has finite length $L_{\mathcal{D}}$ and width $w_{\mathcal{D}}$. The embedding in $t$-th time stamp is generated by $\mathcal{H}$ in the shape of $d_{in}^t \times 1$.

The difference between them and DINVC is that, \textit{We do not take into account the distribution of a given signal.} In fact, the distribution of given images or videos is difficult to approximate, whether it is approached by minimizing ELBO or adversarial training, there is always a certain gap. Flow-based methods or diffusion models suffer from huge computational costs. In contrast, \textit{NeRV system models the distribution of a given video implicitly, through the computation process of decoding. The calculation process per se is regarded as the side information}~\cite{Wyner1976TheRF, Jonschkowski2015PatternsFL}. DINVC constraints the calculation process through the model size.


\textbf{Compare to INR on image}.~\cite{siren} (SIREN) uses sine as a periodic activation function in order to model the high-frequency information of given image.~\cite{fourier} conducts a sinusoidal transformation before input. Their purpose is the reconstruction of fine details.~\cite{Xu_2022_INSP} tries to directly modify an INR without explicit decoding. The difference between them and us is that, the input coordinate-pixel pairs is considered to be \textit{dense} for the INR on image coding. The RGB value of a certain position is closely related to its neighboring positions. But for high-resolution video, the gap between adjacent frames will be much greater in terms of pixel and semantic domain. The situation we focus on in this paper is more similar to that in which we only observe a portion of pixels from a given image. Besides, we have not yet addressed the issue of high-frequency detail recovery in videos in this paper and we would look into it further.

\textbf{Compare to other Kronecker factorization-based models}. \cite{Tahaei2021KroneckerBERTLK} proposes a model compression method using Kronecker decomposition and adopt it in all linear mappings in the Transformer layer~\cite{AttentionIA}. Also, \cite{Jose2017KroneckerRU} decomposes the weight metrices $W$ in RNN into $W = \otimes_{f=0}^{F-1} W_f$. The main difference between them and our method is that \textsc{KFc} also needs to compute the bias term in an efficient manner. We add the lightweight bias calculation via CP decomposition and first adopt the Kronecker decomposition-based operation to the NeRV research.

\textbf{Compare to bilinear feature fusion}. The calculation of \textsc{KFc} seems similar to the multi-head graph attentional layer (GATs)~\cite{gan}, but they are different. The final output of graph attention between $N$ node features $h_j$ and normalized attention coefficients $\alpha$ is calculated as
\[
h_i^{'} = \sigma \left( \sum_{j \in N} \alpha_{ij} \mathbf{W} h_j \right),
\]
where $W$ is the weight metrix. It calculates the global correlation between features in each node. GATs and \textsc{KFc} both have a global receptive field to given features, but another main goal of \textsc{KFc} is to upscale the shape efficiently. The mechanism of \textsc{KFc} is the kronecker decomposition of fully-connected layer and \textsc{KFc} is only formally like GATs or other bilinear models~\cite{Li2016FactorizedBM}. The principal aim of bilinear models is to calculate the relationship between two diverse vectors.

\textbf{Compare to self-attention module}. A self-attention (SA) module~\cite{NonlocalNN, AttentionIA, SwinTH, t2023is} computes the response at a position by attending to all positions, which is similar to \textsc{KFc}. The major defect of SA when adopted in NeRV is that the computational complexity and the space complexity are too high to efficiently compute the global correlations between arbitrary positions, especially the computation between queries and keys for high resolution feature maps. \textsc{KFc} not only captures long-range dependencies but also achieves low-cost rescaling, both of which are significant to NeRV.

\section{Additional Results}
All models used in the additional results are trained under the 3M level size unless otherwise specified.
\subsection{Generalization}
In fact, approximation and generalization are two entirely different topics in deep learning theory~\cite{Nakkiran2019DeepDD, Advani2017HighdimensionalDO, Saxe2018OnTI}, and the relationship between overfitting and generalization of NeRV needs further research. In this paper, we only focus on the approximation ability of models. But we also report on the generalization ability of PNeRV-L through video interpolation. The results are given in Tab.~\ref{int} and PNeRV outperforms most of the baseline methods. Analysis and enhancement of the generalization ability of PNeRV will be investigated in the future.

\begin{table}[h]
    \setlength{\tabcolsep}{3pt}
  \centering
  \setlength{\abovecaptionskip}{5pt}
    \caption{Video interpolation results on 960 $\times$ 1920 UVG in PSNR.}
  \begin{tabular}{l|lllllll|c}
     PSNR         & Beauty & Bospho & Honey & Jockey & Ready & Shake &Yacht &Avg.\\
    \midrule[1.5pt]
    NeRV~\cite{nerv}     &28.05  & 30.04 & 36.99  &20.00 &17.02  & 29.15  &24.50 & 26.54 \\
    E-NeRV~\cite{enerv}  &27.35  & 28.95 & 38.24  &19.39 &16.74  & 30.23  &22.45 & 26.19 \\
    H-NeRV~\cite{HNeRV}  &31.10  & 34.38 & 38.83  &23.82 &20.99  & 32.61  &27.24 & 29.85 \\
    DiffNeRV~\cite{DNeRV}&35.99  & 35.10 & 37.43  &30.61 &24.05  & 35.34  &28.70 & \textbf{32.47} \\
 \midrule
    PNeRV                &33.64  & 34.09 & 39.85  &28.74 &23.12  & 31.49  &27.35 & \underline{31.18} \\
  \end{tabular}
  \label{int}
\end{table}

\subsection{Perceptual Comparison}
We report the LPIPS~\cite{LPIPS} distance of PNeRV (PNeRV-M in 3M and PNeRV-L in 3.3M) comparied with the previous SOTA method (DiffNeRV~\cite{DNeRV} in 3.3M) in Tab.~\ref{LPIPS}. LPIPS is given by a weighted L2 distance between deep features of frames and the AlexNet~\cite{alexnet} which pretrained on ImageNet is used as the deep feature extractor.

Under the same conditions, although the average PSNR of DiffNeRV is only 1.14\% worse, but the average LPIPS is 4.83\% worse than the PNeRV-L. The results suggest that the reconstruction of the proposed methods is better from the point of view of human visual perception. Through the interaction of high-level information, the proposed PNeRV is more likely to benefit downstream video classification, semantic segmentation, object detection, or other semantic-level visual tasks. We would discuss it further in Fig.~\ref{bsm_1} and~\ref{bsm_2}.

\begin{table}[h]\scriptsize
    \tabcolsep=0.08cm
      \setlength{\abovecaptionskip}{5pt}
      \caption{LPIPS comparison on UVG. LPIPS is calculated by trained model which its PSNR is on the right. Avg. is the average LPIPS on whole UVG, the lower the better.}
      \resizebox{!}{0.83cm}{\centering
  \begin{tabular}{l|lllllll|c}
     LPIPS/PSNR         & Beauty & Bospho & Honey & Jockey & Ready & Shake &Yacht &Avg.\\
    \midrule[1.5pt]
    DiffNeRV~\cite{DNeRV}  &0.205/40.00 &0.164/36.58 &0.042/41.97 &0.164/35.75 &0.206/28.16 &0.181/36.53 &0.241/30.43 &0.172/35.63\\
    PNeRV-M                &0.201/39.08 &0.160/35.56 &0.038/42.59 &0.271/31.51 &0.325/25.94 &0.154/37.61 &0.238/30.27 &0.198/34.65\\
    PNeRV-L                &0.210/38.94	&0.132/36.68 &0.037/42.73 &0.177/35.81 &0.211/28.97 &0.146/38.25 &0.230/30.92 &\textbf{0.163}/36.04\\
  \end{tabular}}
  \label{LPIPS}
\end{table}

\begin{figure}[h]
    \centering
	\begin{minipage}{0.45\linewidth}
  \resizebox{!}{0.7\linewidth}{\centering
  \includegraphics[width=1\linewidth]{UVG_msssim6.png}}\\
	\end{minipage}
	\begin{minipage}{0.45\linewidth}
  \resizebox{!}{0.7\linewidth}{\centering
  \includegraphics[width=1\linewidth]{train1.png}}
  \\ 
	\end{minipage}
\vspace{-0.05in}\caption{Left: Comparison of video compression on UVG in MS-SSIM. Right: A few examples of training failures for various NeRV systems on Parkour.}\label{train}
\end{figure}

\begin{figure}[t]
  \centering
  \includegraphics[width=1\linewidth]{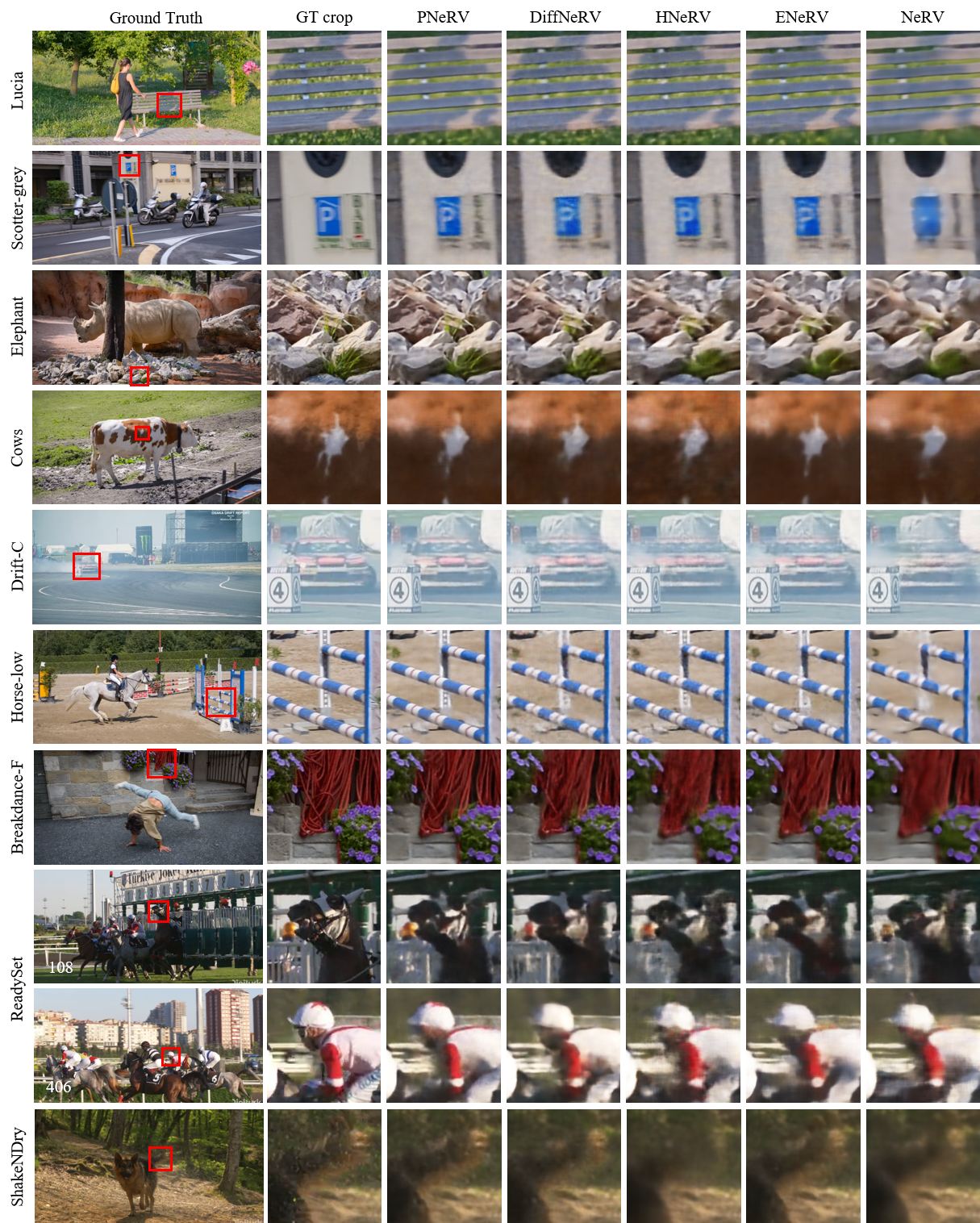}\\
  \vspace{-0.05in}\caption{Visual comparison examples on various videos.}
  \label{add_davis}
\end{figure}
\subsection{Discussion on Video Compression}
The video compression comparison of PNeRV with other NeRV models in MS-SSIM is shown on the left of Fig.~\ref{train}. It is worth noting that, ALL NeRV methods suffer from dramatic training difficulty under large parameters like 5M or more. Models will fall into local minima result in trivial outputs. This phenomenon seriously hinders the performance of all NeRV methods when Bpp gets larger. Examples of failure training are given to the right of Fig.~\ref{train}, in which the models are trained under the same conditions.

\subsection{More Qualitative Comparisons}
We show some more examples of qualitative comparisons between different models.

Shown in Fig.~\ref{add_davis}, the results of PNeRV are more smooth and less noisy. Such as ``Lucia'' and ``Horse-low'', PNeRV pays more attention to the geometric pattern of main objects and ignores those high-frequency details of the background scene. Other baseline methods could not reconstruct frames in such semantic level. Due to the lack of high-level information guidance and global receptive field, baseline methods are hard to reasonably allocate model weights to more important objects, e.g., red waterpipe in ``Breakdance-flare'' and patterns in ``Cows''.

\begin{figure}[t]
  \centering
  \includegraphics[width=1\linewidth]{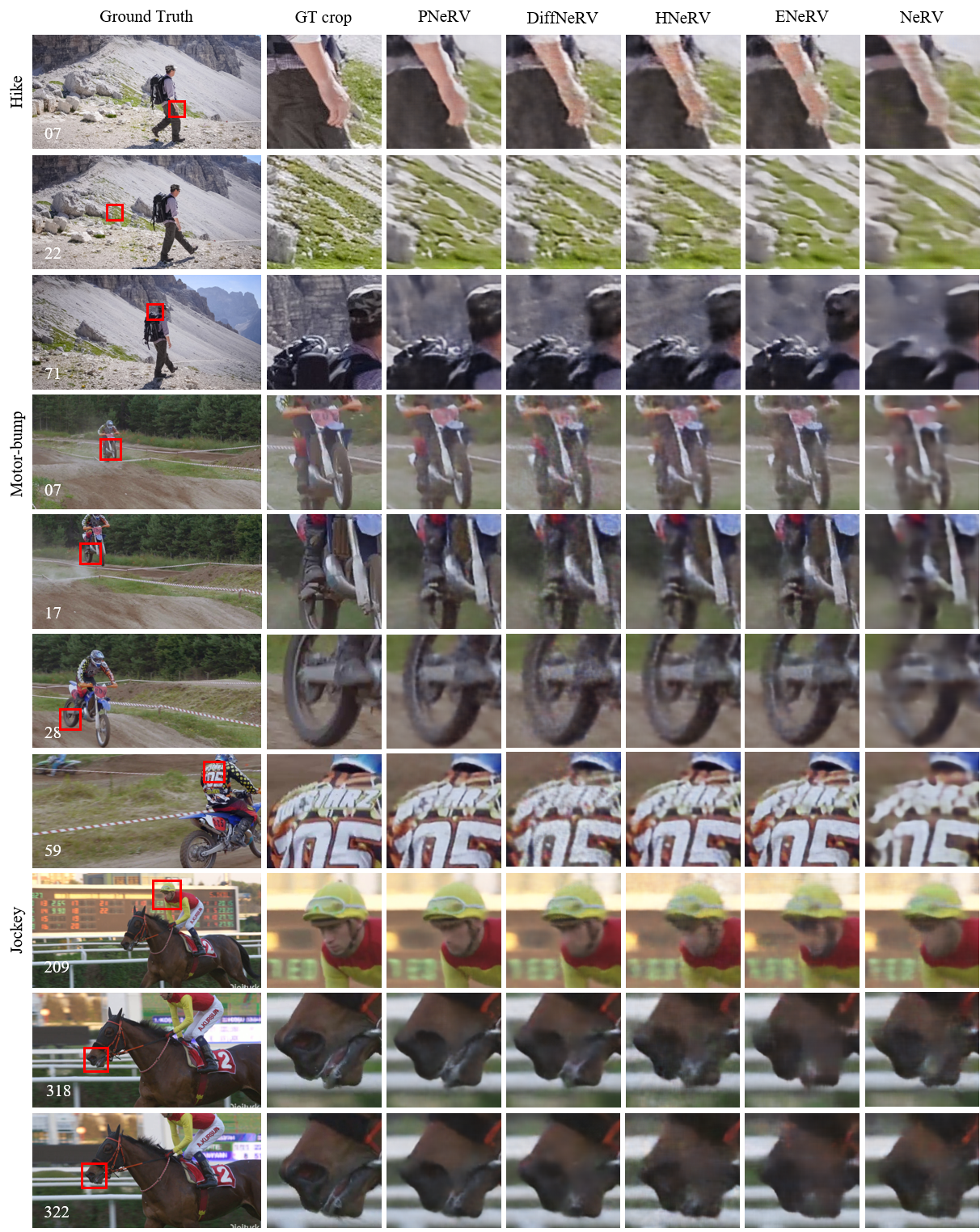}\\
  \vspace{-0.05in}\caption{Visual comparison examples on the same video. Corresponding time stamps are shown in the bottom left.}
  \label{same_davis}
\end{figure}

Shown in Fig.~\ref{same_davis}, the comparison in different time stamp of same video indicates some specific common issues of different models.
Overlapping and noisy patterns are occurred in the results of DiffNeRV~\cite{DNeRV} and HNeRV~\cite{HNeRV} such as grass and hands in ``Hike''. ENeRV~\cite{enerv} and NeRV~\cite{nerv} often result in color deviation and blurring, e.g. backpack in ``Hike'' and motor in ``motor-bump''. PNeRV achieves a balance between preserving details and maintaining the semantic consistency of spatial patterns. The reconstruction of DiffNeRV, which also uses difference of frames as input, is unbiased when recovering the fine details. However, in fact, human attention to visual elements with different semantics should be different. Improving the reconstruction results through high-level information is one of PNeRV's pursuits.

\section{Additional ablation studies}
\begin{figure}[t]
  \centering
  \includegraphics[width=1\linewidth]{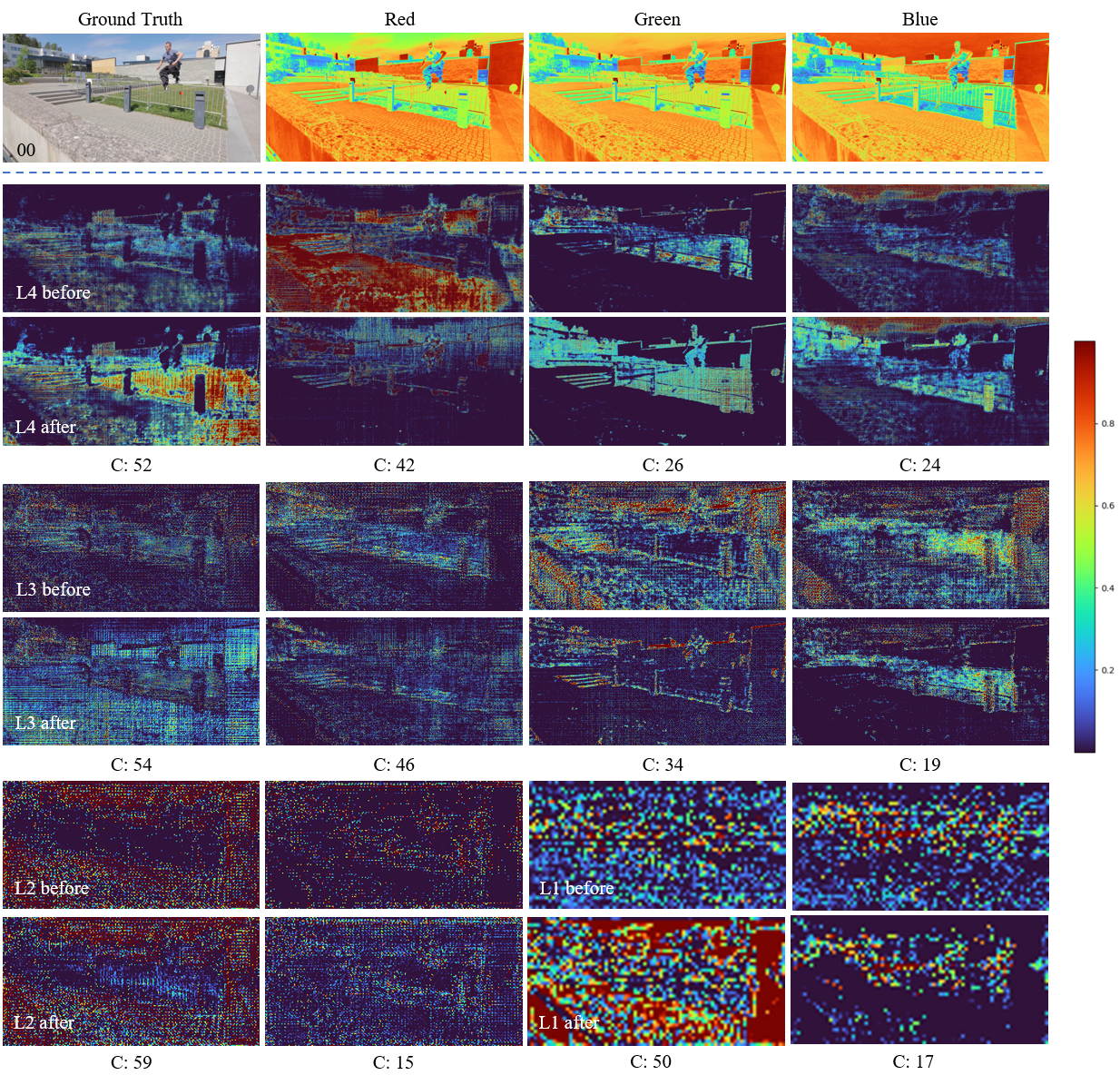}\\
  \vspace{-0.05in}\caption{Visualization examples of feature maps in different layers.``C'' refers to the channel number and ``L'' is the layer number.}
  \label{bsm_1}
\end{figure}

\begin{figure}[t]
  \centering
  \includegraphics[width=1\linewidth]{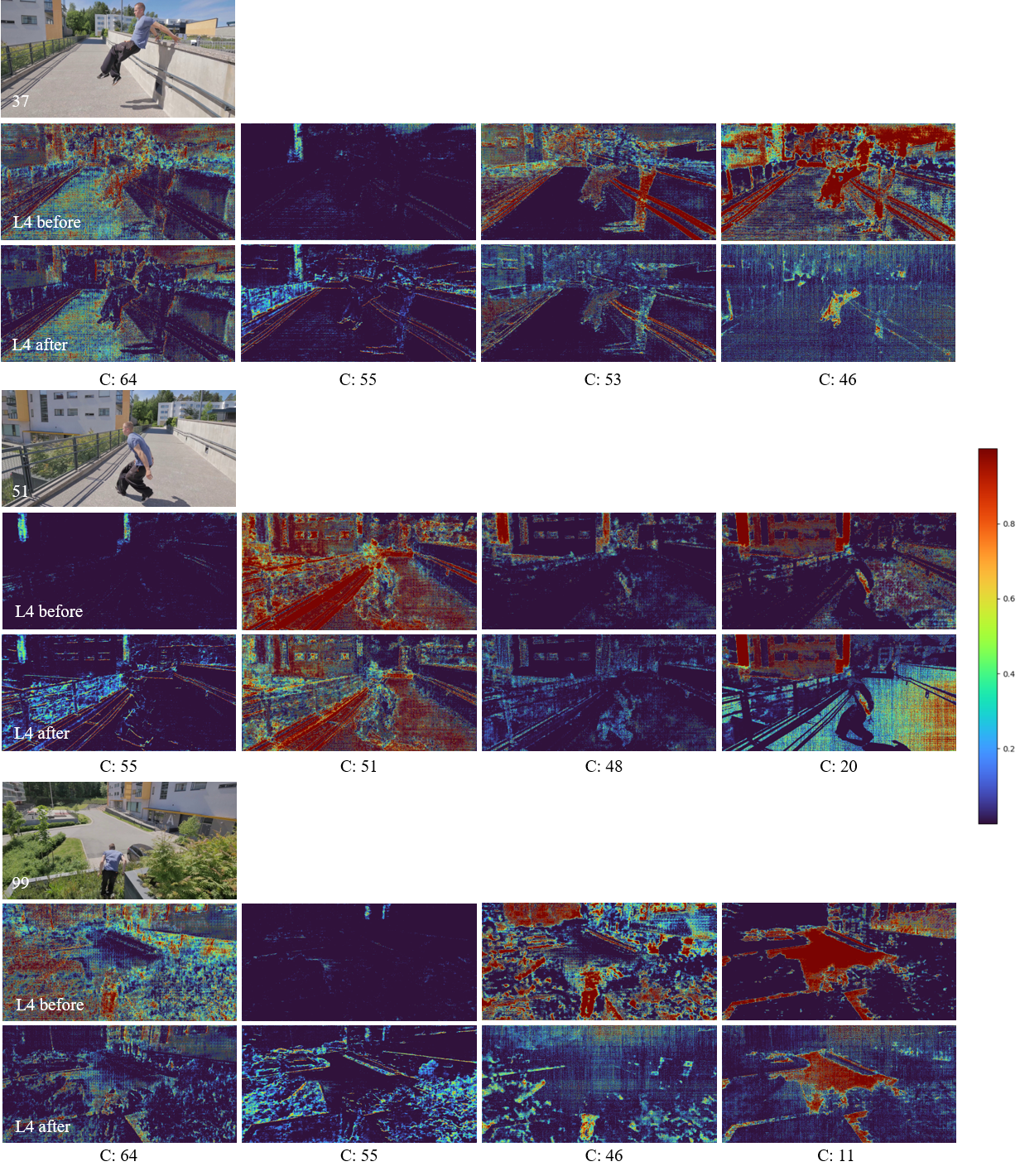}\\
  \vspace{-0.05in}\caption{Visualization examples of feature maps for different frames.``C'' refers to the channel number and ``L'' is the layer number.}
  \label{bsm_2}
\end{figure}

We visualize the feature maps in PNeRV-L trained on ``Parkour'' as examples to demonstrate the efficacy of hierarchical information merging using \textsc{KFC} and BSM. Those feature maps shown in Fig.~\ref{bsm_1} are on different channels and layers using the same frame as input. Those in Fig.~\ref{bsm_2} are all from $4$-th layer but using different frames as input. The feature maps from $4$-th layer are in $480 \times 960$ and the original frames are in $960 \times 1920$. For each lower layer, the height and width are divided by 2 compared to the upper layer. ``Before'' and ``After'' means those maps are before passing through BSM or after.

Fig.~\ref{bsm_1} indicates that, the coarse features are refined by BSM. Different channels' features react differently to various aspects of video frames, such as color, geometric shape, texture, brightness, motion, etc. Before passing BSM, the vanilla features are mix of various elements, but BSM decouples them and distinguish their specific effect.

Also, for the defective feature maps, BSM could add additional details or balance the focus of the reconstruction on various area in frames. These phenomenons occur universally in $4$-th layer which prepareing for fine-grained reconstruction, as shown in the Fig.~\ref{bsm_2}.

\section{Proof of Theorem~\ref{param}}\label{app_a}

Before we start, we will recall the setup and demonstrate the lemmas.
\begin{definition}\label{string}
A function $g : \mathbb{R}^{d_{in}} \to \mathbb{R}^{d_{out}}$ is a max-min string of length $L \geq 1$ on $d_{in}$ input variables and $d_{out}$ output variables if there exist affine functions ${\ell}_1, \cdots ,{\ell}_L: \mathbb{R}^{d_{in}} \to \mathbb{R}^{d_{out}}$ such that
\[
g = \sigma_{L-1} ( \ell_L, \sigma_{L-2} \left(  \ell_{L-1},\cdots , \sigma_2 \left( \ell_3, \sigma_1\left( \ell_1, \ell_2\right)\right) \cdots \right).
\]
\end{definition}

\begin{definition}\label{w}
The width $w$ of a feed-forward network is the maximal width $w_l$ among all $L$ hidden layers, $w = \max( \{ w_l \}^L_{l=1}  )$ .
\end{definition}

The definition of max-min string and DMoC are first introduced in~\cite{UniversalFA} and~\cite{ApproximatingCF}. We introduce two lemmas which given as Propositions 2 and 3 in~\cite{ApproximatingCF}.
\begin{lemma}\label{L1}
For every max-min string $g$ on $d_{in}$ input variables and $d_{out}$ output variables with length $L$ and every compact $K \subseteq \mathbb{R}^{d_{in}}$, there exists a \textsc{ReLU} net $\mathcal{N}$ with input dimension $d_{in}$, hidden layer width $d_{in}+d_{out}$, and depth $L$ that computes $x \mapsto g(x)$ for every $x \in K$.
\end{lemma}

\begin{lemma}\label{L2}
For every compact $K \subseteq \mathbb{R}^{d_{in}}$, any continuous $f : K \to \mathbb{R}^{d_{out}}$ and each $\epsilon \geq 0$ there exists a max-min string $g$ on $d_{in}$ input variables and $d_{out}$ output variables with length
\[
\left( \frac{ \mathop{O}
    \left(diam  \left(K \right)
    \right) }{\omega_f^{-1} \left( \epsilon \right) }
\right)^{d_{in}+1},
\]
for which
\[
\Vert f-g \Vert_{C^0(K)} \leq \epsilon.
\]
\end{lemma}


\begin{lemma}\label{L3}
For any $p \in [1, \infty )$, \textsc{ReLU} nets of width $w$ are dense in $L^P (\mathbb{R}^{d_{in}}, \mathbb{R}^{d_{out}})$ if and only if $w \geq \max\{d_{in} + 1, d_{out}\}$.
\end{lemma}

\begin{theorem}\label{param}
For a NeRV system $\mathcal{G}$ where its decoder $\mathcal{D}$ calculating in serial to $\epsilon$-approximate a given video $\mathcal{V}$ which is implicitly characterized by a certain unknown L-lipschitz continuous function $\mathcal{F}_{\mathcal{V}}: K \to \mathbb{R}^{d_{out}}$ where $K \subseteq \mathbb{R}^{d_{in}}$ is a compact set, then the upper bound of the minimal parameter quantity $\mathsf{Param}(\mathcal{D})$ for $\mathcal{G}$ is given as\vspace{-0.05in}
\[
 \mathsf{Param}_{\min}(\mathcal{D})\leq d_{out}^2 \left( \frac{ \mathop{O}
    \left(diam  \left(K \right)
    \right) }{\omega_f^{-1} \left( \epsilon \right) }
\right)^{d_{in}+1}.
\]
\end{theorem}
\begin{proof}
From Lemma~\ref{L2}, the implicit function $\mathcal{F}_{\mathcal{V}}$ which represents the video $\mathcal{V}$ can be approximated by one max-min string $g$ which defined in~\ref{string}. It is worth mentioning that $\mathcal{F}_{\mathcal{V}}$ is supposed to be continuous because video can be considered as a slice of Minkowski space in real life. The length of the max-min string $g$ is given by Lemma~\ref{L2}. According to Lemma~\ref{L2}, there exists a \textsc{ReLU} net $\mathcal{N}_g$ with same input and output dimensions that fit the max-min string. So, the minimal parameters of $\mathcal{N}_g$, also the sum of weights for each affine layer, is
\[
\mathbf{Param} =\sum_{l=1}^{L} w_l w_{l-1},
\]
where $L$ is given in Lemma~\ref{L2}. Further the minimal width $w_{\min}$ is estimated as $\max\{d_{in}+1, d_{out} \}$ by~\cite{MinimumWF} (Lemma~\ref{L3}), thus the minimal parameters of $\mathcal{N}_g$ under certain error is no longer than
\[
\mathbf{Param}_{\min} \leq w_{\min}^2 \left( \frac{ \mathop{O}
    \left(diam  \left(K \right)
    \right) }{\omega_f^{-1} \left( \epsilon \right) }
\right)^{d_{in}+1} = d_{out}^2 \left( \frac{ \mathop{O}
    \left(diam  \left(K \right)
    \right) }{\omega_f^{-1} \left( \epsilon \right) }
\right)^{d_{in}+1},
\]
where $w_{\min}= d_{out}$ for video $\mathcal{V}: \mathbb{N}\to \mathbb{R}^{d_{out}}$. Equality is reached when each hidden layer width reaches the upper bound of minimal width, $w_l = w = w_{min}$.

\end{proof}

\section{Pseudocode}
The proposed PNeRV framework is easy to implement based on the commonly used layers. We provide a PyTorch-like pseudocode of single layer in PNeRV as follows.

\begin{python}\scriptsize
import torch
import torch.nn as nn
class KFc(nn.Module):
	def __init__(self, in_height, in_width, out_height, out_width, channels):
		super().__init__()
		self.in_h = in_height
		self.in_w = in_width
		self.out_h = out_height
		self.out_w = out_width
		self.c = channels

		self.w_L_ = torch.zeros(self.c, self.out_h, self.in_h)
		self.w_R_ = torch.zeros(self.c, self.in_w, self.out_w)
		nn.init.kaiming_normal_(self.w_L_, mode='fan_out', nonlinearity='relu')
		nn.init.kaiming_normal_(self.w_R_, mode='fan_out', nonlinearity='relu')
		self.w_L = nn.Parameter(self.w_L_)
		self.w_R = nn.Parameter(self.w_R_)

		self.b_h = nn.Parameter(torch.zeros(self.out_h, 1))
		self.b_w = nn.Parameter(torch.zeros(1, self.out_w))
		self.b_c = nn.Parameter(torch.zeros(self.c, 1))

	def forward(self, x):
		b_ = self.b_h @ self.b_w
		b__ = b_.reshape(1, self.out_h*self.out_w)
		_ = self.b_c @ b__
		__ = _.reshape(self.c, self.out_h, self.out_w)
		b = __.repeat(self.in_b, 1, 1, 1)
		_ = torch.matmul(self.w_L, x)
		return torch.matmul(_, self.w_R) + b

def PNeRV_one_layer(embedding, input):
    c,h,w = embedding.shape
    C,H,W = input.shape
    #KFc layer
    pym = KFc(h, w, H, W, c)(embedding)
    pym = nn.BatchNorm2d(c, track_running_stats=True)(pym)
    pym = nn.GELU()(pym)
    #BSM layer
    memory = nn.Conv2d(C, C, 3, 1, 1)(input)
    kownledge = nn.Conv2d(c, C, 3, 1, 1)(pym)
    _ = torch.relu(memory + kownledge)
    decision = torch.sigmoid(nn.Conv2d(C, C, 3, 1, 1)(_))
    output = decision * knowledge + (1- decision) * input
    return basic_NeRV_block(output)
\end{python}

{\small
\bibliographystyle{unsrt}
\bibliography{n23}
}
